\documentclass[lettersize,journal]{IEEEtran}
\usepackage{amsmath,amssymb,amsfonts}
\usepackage{algorithmic}
\usepackage{algorithm}
\usepackage{array}
\usepackage{url}

\usepackage{float}
\usepackage{subfigure,multirow,bm,stfloats}
\usepackage{textcomp}
\usepackage{amsthm}

\usepackage{graphicx}
\usepackage[table,xcdraw]{xcolor}
\begin{document}

\title{Hyperspectral Image Segmentation based on Graph Processing over Multilayer Networks}

\author{Songyang Zhang, Qinwen Deng, and Zhi Ding,~\IEEEmembership{Fellow,~IEEE,}
        % <-this % stops a space

\thanks{S. Zhang, Q. Deng, and Z. Ding are with Department of Electrical and Computer Engineering, University of California, Davis, CA, 95616. (E-mail: sydzhang@ucdavis.edu, mrdeng@ucdavis.edu, and zding@ucdavis.edu)}
\thanks{This material is based upon work supported by the National Science Foundation under Grants No.  2029848 and 2002937.}
}



\maketitle

\begin{abstract}
Hyperspectral imaging is an important
sensing technology with broad applications and impact in
areas including
environmental science, weather, and 
geo/space exploration. One important task 
of  hyperspectral image (HSI)
processing is the extraction of spectral-spatial features. 
Leveraging on the recent-developed graph signal processing over multilayer networks (M-GSP), this work proposes several approaches to HSI segmentation based on M-GSP feature extraction. To capture joint spectral-spatial information, we first customize a tensor-based multilayer network (MLN) model for HSI, and define a MLN singular space for feature extraction. We then develop an unsupervised HSI segmentation method by utilizing
MLN spectral clustering. Regrouping 
HSI pixels via MLN-based clustering, we further propose a semi-supervised HSI classification based on  multi-resolution fusions of superpixels. Our experimental results demonstrate the strength of M-GSP in HSI processing and spectral-spatial information extraction.
\end{abstract}


\section{Introduction}
\begin{figure*}[t]
	\centering
	\includegraphics[width=7in]{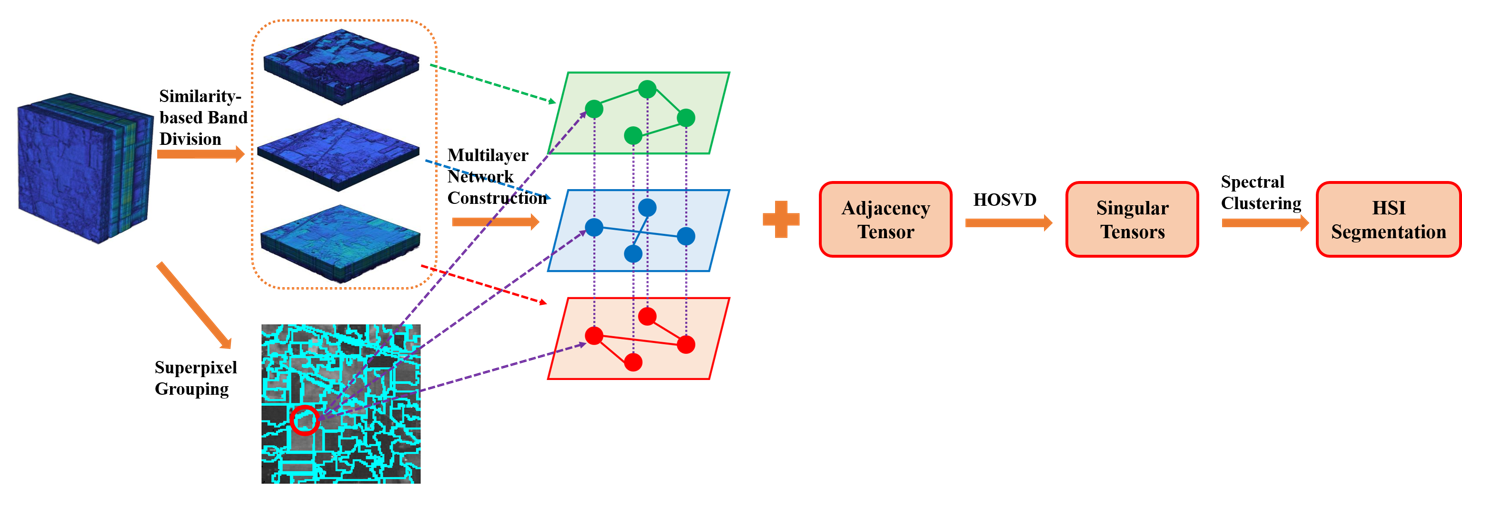}
	\caption{Scheme of MLN-based Unsupervised HSI Segmentation.}
	\label{unsupervised}
\end{figure*}

\begin{figure*}[t]
	\centering
	\includegraphics[width=7.2in]{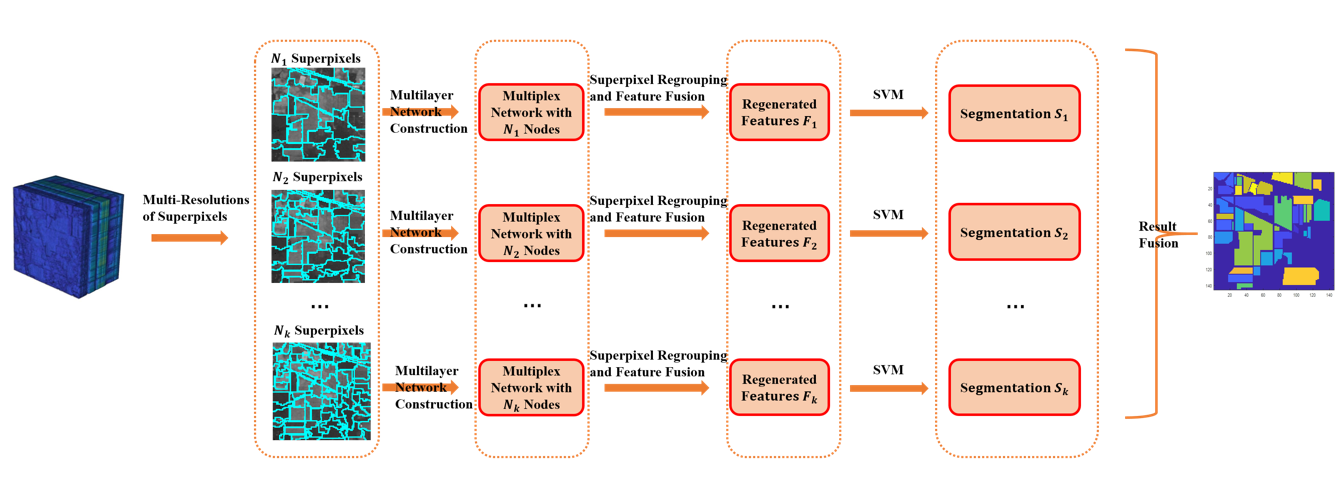}
	\caption{Scheme of MLN-based Semi-supervised HSI Segmentation.}
	\label{supervised}
\end{figure*}

\IEEEPARstart{H}{yperspectral} imaging is an analytical technique
operating on images at different wavelengths for given geographical areas \cite{A1,c1,c2}. Exploiting a wealth of
spectral-spatial information, hyperspectral images (HSIs) have seen broad applications in areas such as urban mapping, environment management, crop analysis and food safety inspection \cite{c3,c4,c5}.
Many such applications often require to label
each spatial position shown as an image 
pixel.  Thus, hyperspectral image segmentation 
has emerged as an important field in HSI analysis \cite{c6}. 
Given the corresponding spectral feature vectors for each position
in HSI, HSI segmentation aims to partition image pixels into 
different feature groups (or regions). Typical types of HSI segmentation include: 1) supervised (semi-supervised) classification given the labels of a training sample set, and 2) unsupervised segmentation without prior knowledge on the labels of pixels. In this work, we investigate the application of multilayer network graph signal processing (M-GSP) in both unsupervised clustering and semi-supervised classification as conceptually  
illustrated in Fig. \ref{unsupervised} and Fig. \ref{supervised}, respectively.

Supervised HSI classification approaches form an important category of HSI analysis in remote sensing and have shown strong performance on standard datasets \cite{c7}. One typical approach is to first reduce the feature dimension before applying a classifier, such as support vector machines (SVM) for segmentation \cite{c8,c9}. 
To remove redundancy and to reject noise across the spectral bands, different feature selection \cite{c10} and feature extraction \cite{c3} methods have shown successes. Among them,
we see familiar concepts that leverage 
principal component analysis (PCA) \cite{c11,c12,c13,c14,c15}, independent component analysis (ICA) \cite{c16}, and manifold learning \cite{c17,c18}, among others. With the recent advances in machine learning, newly proposals such as deep learning \cite{c19}, active learning \cite{c20}, and tensor learning \cite{c21}, have also 
demonstrated robust performance of HSI classification. Despite their successes,
HSI classification algorithms require a significant number of 
high quality samples for model optimization and can be costly in
many real-life scenarios \cite{c22}. On the other hand, unsupervised segmentation is also an important task in HSI analysis and has attracted significant coverage recently. Basic Centroid-based clustering methods, such as $k$-means \cite{c23} and fuzzy $c$-means \cite{c24}, aim to minimize intra-cluster distance of 
samples within the clusters. Other clustering methods, including density-based methods \cite{c25} and biological clustering methods \cite{c26}, have also been 
adopted for unsupervised HSI segmentation.

In addition to the aforementioned approaches, graph-based methods have been gaining
popularity in both supervised (semi-supervised) and unsupervised HSI segmentation, 
owing to their power in revealing underlying structures among different pixels. Representing HSIs as graphs, spectral clustering \cite{c27,c22} can be developed for the unsupervised HSI segmentation. Compared to the centroid-based clustering and density-based methods,  graph-based spectral clustering can 
provide a more general similarity model for pixels and perform well in
unsupervised HSI segmentation. Similarly, graph convolutional network (GCN) \cite{c29,c30,c31,c32} has also demonstrated its strength in 
(semi-) supervised HSI segmentation, especially under the condition
of limited ground-truth hyperspectral sets \cite{c33}, by exploring the underlying structures among the pixels.

\begin{figure}[t]
	\centering
	\subfigure[]{
		\label{mln1}
		\includegraphics[height=3.5cm]{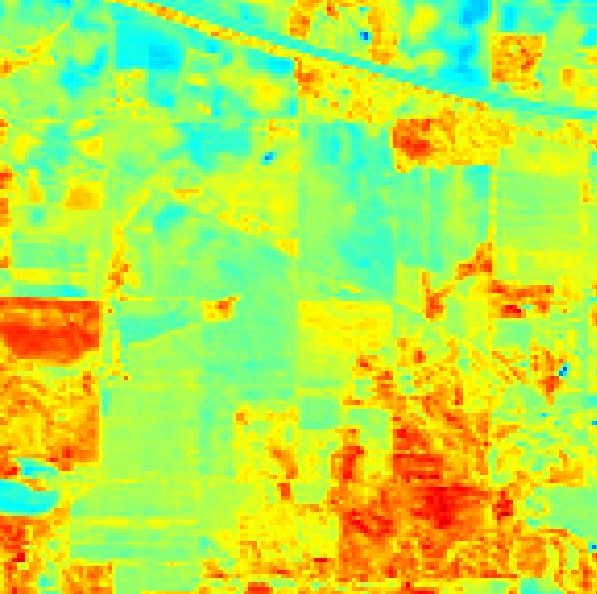}}
	\hfill
	\subfigure[]{
		\label{mlp1}
		\includegraphics[height=3.5cm]{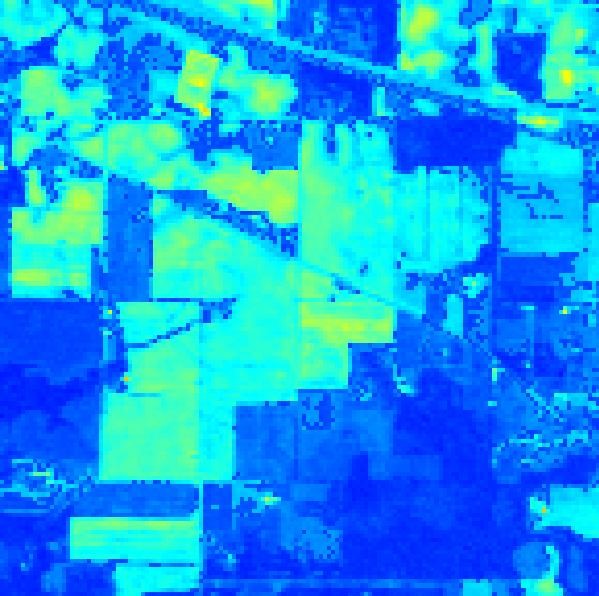}}
	\caption{Examples of Volumes for Different Bands in the Indian Pines Dataset.}
	\label{ex_mln1}
\end{figure}

Despite these reported successes, most of the graph-based methods  focus primarily
on spatial geometry and consider a rather stationary
graph connections for all spectral bands. Such limitation fails to
explore different spectrum features of each individual band. For example, Fig. \ref{ex_mln1} shows that different bands may display different distributions of the pixel volumes, which could mean different graph structures. To compensate, some works would represent each band with an individual graph while neglecting the inter-frame correlations \cite{c34}. For these reasons, the unresolved challenge is to 
understand how to capture such heterogeneous spectral-spatial structure in
an integrative manner jointly instead of individually. 

In addition to the limitations of geometric models in general, traditional graph-based methods may also suffer from the large number of HSI pixels to process \cite{c22}. Generally, superpixels are constructed to control complexity. However, to achieve a better performance,  one needs to determine the resolution of superpixels in advance and also decide how to
form superpixels. In \cite{c15}, different resolutions of superpixels are combined after the individual processing for each of them, which presents the ensuing challenges on
the efficient fusion of the segmented results from different resolutions.

In view of the aforementioned challenges, this work introduces 
a new approach of graph signal processing over multilayer networks (M-GSP) to
HSI processing. M-GSP \cite{c35,c36} is a tensor-based framework that generalizes traditional graph signal processing (GSP) \cite{c37} to
process the heterogeneous graph structures across different layers
of graphs. Since the spatial positions (pixels) are the same for all hyperspectral
bands, we can model HSI data as a multilayer network with the same number of nodes in each layer (also known as multiplex network). The superpixels serve as nodes and each spectrum frame forms one layer as shown in Fig. \ref{unsupervised}. Next,
we develop M-GSP spectral analysis to extract features for HSI segmentation. 
In this work, we investigate the applications of M-GSP in the feature extraction for both unsupervised and (semi-) supervised HSI segmentation. We summarize our contributions in this work as follows:

\begin{itemize}
	\item We propose a multilayer network (MLN) model, together with an alternative singular space for the HSI datasets.
	\item We develop an unsupervised HSI segmentation method based on the M-GSP spectral clustering.
	\item We develop an MLN-based method for semi-supervised HSI segmentation, which combines multi-resolution information. 
	\item We further propose several novel schemes for decision fusion of the results from different resolutions of superpixels.
\end{itemize}
We test all the algorithms in the standard \textit{Indian Pines} dataset, \textit{Pavia University} dataset and \textit{Salinas} dataset. The experimental results 
demonstrate the strength of M-GSP in representing the spectral-spatial structures in
HSI, and the efficiency of the proposed HSI segmentation algorithms.

We organize the rest of this manuscript as follows. Starting with a brief introduction of M-GSP in Section \ref{mgsp}, we introduce the multilayer network construction for HSI datasets together with the introduction of unsupervised HSI segmentation based on M-GSP spectral clustering in Section \ref{unsup}. Next, we introduce the M-GSP framework of semi-supervised HSI classification in Section \ref{sup}, where we also propose several methods for the decision fusion of different resolutions. In Section \ref{exp}, we present details on the experiments and results of the proposed methods in several standard datasets. Finally, we summarize our works in Section \ref{con}.

\section{Fundamentals of M-GSP}\label{mgsp}
In this section, we briefly introduce the core concepts of M-GSP used in this work. 

A multilayer network $\mathcal{M}$ with $M$ layers and $N$ nodes in each layer can be viewed as
projecting $N$ virtual entities (nodes) onto $M$ layers. The resulting $M$
layers form a multilayer ``network'' $\mathcal{M}$ that
be represented by a fourth-order adjacency tensor $\mathbf{A}\in \mathbb{R}^M\times \mathbb{R}^N\times 
\mathbb{R}^M \times \mathbb{R}^N$ defined as
\begin{equation}\label{adj}
\mathbf{A}=(A_{\alpha i \beta j}), \quad 1\leq \alpha,\beta\leq M, 1\leq i,j\leq N,
\end{equation}
where each entry $A_{\alpha i \beta j}$ of $\mathbf{A}$ indicates the edge (link) intensity 
between entity $j$'s projection in layer $\beta$ and entity $i$'s projection in layer $\alpha$ \cite{c38}. 
Generally, we use bold uppercase letters
to denote tensors, i.e.,  $\mathbf{A}\in\mathbb{R}^{I_1}\times \cdots\times \mathbb{R}^{I_Q}$ represents a $Q$th-order tensor with $I_k$ being the dimension of the $k$th order.
$A_{{i_1}\cdots i_Q}$ denotes the entry of $\mathbf{A}$ 
in position $(i_1,i_2,\cdots,i_Q)$ with $1\leq i_k\leq I_k$. 
%For tensor $\mathbf{A}_f$ with an index $f$, we write its entries
%as $[A_f]_{{i_1}\cdots i_N}$.
In addition to adjacency tensor, Laplacian tensor can be also defined similar to that in normal graphs. Interested reader could refer to \cite{c35} for more details.

Since we construct the superpixels from all spectrum images in a HSI dataset, each superpixel can be viewed as an entity to be projected into different spectrum bands. 
We can then represent the spectral-spatial structure intuitively via the adjacency tensor in Eq. (\ref{adj}). More details of the multilayer construction 
will be discussed in Section \ref{unsup}.

With the adjacency
tensor $\mathbf{A}\in\mathbb{R}^M\times \mathbb{R}^N\times 
\mathbb{R}^M \times \mathbb{R}^N$, it can be decomposed via higher-order singular value decomposition (HOSVD) \cite{c39} as
\begin{equation}\label{decomposeS}
\mathbf{A}\approx \mathbf{S}\times_1 \mathbf{U}^{(1)}\times_2 \mathbf{U}^{(2)}\times_3 \mathbf{U}^{(3)}\times_4 \mathbf{U}^{(4)},
\end{equation}
where $\times_n$ is the n-mode product \cite{c40} and $\mathbf{U}^{(n)}=[\mathbf{u}^{(n)}_1\quad \mathbf{u}^{(n)}_2\quad\cdots\quad \mathbf{u}^{(n)}_{I_n}]$ is a unitary $(I_n\times I_n)$ matrix, with $I_1=I_3=M$ and $I_2=I_4=N$.
$\mathbf{S}$ is a $(I_1\times I_2\times I_3 \times I_4)$-tensor of which the subtensor $\mathbf{S}_{i_n}$ obtained by freezing
the $n$th index to $\alpha$:
\begin{itemize}
	\item $<\mathbf{S}_{i_n=\alpha},\mathbf{S}_{i_n=\beta}>=0$ where $\alpha\neq\beta$.
	\item $||\mathbf{S}_{i_n=1}||\geq||\mathbf{S}_{i_n=2}||\geq\cdots\geq ||\mathbf{S}_{i_n=I_n}||\geq 0$.
\end{itemize}

The Frobenius-norms $\sigma_i^{(n)}=||\mathbf{S}_{i_n=i}||$ is the $n$-mode singular value, with corresponding 
singular vectors in $\mathbf{U}^{(i)}$. Since the representing tensor is partial symmetric
in the undirected MLN, there are two modes of singular spectrum, i.e., $(\gamma_\alpha, \mathbf{f}_\alpha)$ for mode $1,3$, and $(\sigma_i,\mathbf{e}_i)$ for mode $2,4$. 
More specifically, $\mathbf{U}^{(1)}=\mathbf{U}^{(3)}=(\mathbf{f}_\alpha)$ characterizes the features of layers and $\mathbf{U}^{(2)}=\mathbf{U}^{(4)}=(\mathbf{e}_i)$ characterizes the entities.

If the singular vectors are included in $\mathbf{F}_s=[\mathbf{f}_1\cdots\mathbf{f}_M]\in\mathbb{R}^M\times 
\mathbb{R}^M$ and $\mathbf{E}_s=[\mathbf{e}_1\cdots\mathbf{e}_N]\in\mathbb{R}^N\times \mathbb{R}^N$, the MLN singular transform (M-GST) for a MLN signal $\mathbf{s}\in\mathbb{R}^{M\times N}$ can be defined as
\begin{equation}
\check{\mathbf{s}}_L=\mathbf{F}_s^{\mathrm{T}}\mathbf{s}\mathbf{E}_s\in\mathbb{R}^M\times \mathbb{R}^N.
\end{equation}

Here, we mainly focus on fundamentals of singular analysis of the undirected multilayer networks. 
For more details on other concepts, such as MLN Fourier transform, M-GSP filter design, sampling theory and stationary process, readers are referred to \cite{c35,c36}.

\section{Unsupervised HSI Segmentation based on M-GSP Spectral Clustering}\label{unsup}
In this section, we introduce the construction of MLN models for HSI datasets, before proposing
an unsupervised segmentation approaches based on M-GSP spectral clustering.
\subsection{Superpixel Segmentation for HSI}
Before venturing into the M-GSP analysis, we first introduce the superpixel segmentation for HSI. In traditional graph-based HSI analysis, image
pixels act as nodes and their pair-wise distances are calculated to form a graph \cite{c22}. However, given a large number of pixels, it becomes inefficient and sometimes impossible to implement full graph-based analysis for pixel-based HSIs. Practically, since pixels within a small region may share similar features, grouping neighboring pixels into superpixels could be a more practical way for graph construction.

In general, a suitable superpixel segmentation algorithm for HSI classification should exhibit low computation complexity and accurate
detection of the object boundaries \cite{c15}. One category of superpixel segmentation in HSI uses image features such brightness, color and texture cues, to estimate the location of segment boundaries. In \cite{c41}, a superpixel estimation is developed by adopting the ultrametric contour map (UCM) approaches to the hyperspectral volumes. In \cite{c42}, the band smoothness is jointly considered with general features to group pixels. Graph-based segmentation approaches are also common
in superpixel segmentation \cite{c43}. In \cite{c44}, an eigen-based solution to  normalized cuts (NCuts) is used for superpixel group. However, such eigen-based methods tend to suffer from time-consuming graph construction and matrix decomposition. 

For more efficient superpixel segmentations, our work in this manuscript
considers the entropy rate superpixel segmentation (ERS) suggested by,
e.g., \cite{c15}. In ERS \cite{c45}, a dataset is modeled as a graph $\mathcal{G}=\{\mathcal{V}, \mathcal{E}\}$, in which the pixels serve as the nodes $\mathcal{V}$ and their pairwise similarities are represented by edges $\mathcal{E}$. Next, a subgraph $\mathcal{A}=\{\mathcal{V},\mathcal{L}\}$ is formed by choosing a subset of edges $\mathcal{L}\subseteq\mathcal{E}$, such that $\mathcal{A}$ consists of fewer connected components. To obtain such a subgraph, the problem can be formulated as
\begin{align}\label{obj}
	&\mathcal{L}^*=\arg_{\mathcal{L}} \max{Tr\{H(\mathcal{L})+\alpha T(\mathcal{L})\}}\\
	&s.t. \quad \mathcal{L}\subseteq\mathcal{E},
\end{align}
in which entropy rate term $H(\mathcal{L})$ favors more
compact clusters, and regularizing term $T(\mathcal{L})$ punishes 
large cluster size. Based on the objective function of Eq. (\ref{obj}), a greedy algorithm can be implemented to solve the problem \cite{c46}.

\subsection{Multilayer Network Construction for HSI Datasets}
With superpixel-represented HSI, we now begin multilayer network construction. Consider Fig. \ref{unsupervised}. An HSI $\mathbf{X}\in\mathbb{R}^{K}\times\mathbb{R}^{N}$, containing $K$ spectrum frames and $N$ superpixels, can be modeled by a multilayer network with $M$ layers and $N$ nodes in each layer. Specifically, the MLN consists of the following attributes
\begin{itemize}
	\item \textit{Layers}: To construct a MLN, we define layers based on the spectrum bands. Since different spectrum frames may share similar features, we first divide the bands into $M$ clusters, i.e., $\mathbf{X}_i\in\mathbb{R}^{{K}_i}\times \mathbb{R}^{N}$, $i=1,\cdots,M$ and $\sum_{i=1}^{M}K_i=K$. Next, each cluster serves as one layer in the multilayer network. Various clustering methods can generate features $\mathbf{X}_i$ for layer $i$. For example, one can divide spectrum band based on 
	a range of wavelength. To capture correlation across different bands more efficiently, the $k$-means clustering is applied for band division.
	
	\item \textit{Nodes}: In M-GSP based HSI processing, the superpixels act as  virtual entities. By projecting $N$ superpixels into $M$ layers, we form a multilayer network with $M$ layers and $N$ nodes in each layer. 
	We define MLN signals as the divided attributes of each superpixel, i.e., $\mathbf{X}_{i,j}\in\mathbb{R}^{K_i}$ for the superpixel $j$'s projected node in layer $i$.
	\item \textit{Interlayer connections}: For interlayer connections, each projected node is connected to its counterparts in other layers, i.e., fully connected for all the projected nodes of the same superpixel. As a result, we obtain corresponding entries of the adjacency tensor $\mathbf{A}\in \mathbb{R}^M\times \mathbb{R}^N\times 
	\mathbb{R}^M \times \mathbb{R}^N$ for the superpixel $i$ defined as
	\begin{align}\label{inter}
		A_{\alpha i \beta i}=\left\{\begin{aligned}
		1, & \quad \alpha\neq\beta;\\
		0, & \quad otherwise,
		\end{aligned}
		\right.
	\end{align}
	where each entry indicates link presence.
	\item \textit{Intralayer connections}: For the intralayer connections, we calculate the weights between the projected nodes of entities $i$ and $j$ in layer $\alpha$ based on the localized Gaussian distance as follows:
	\begin{align}\label{intra}
	A_{\alpha i \alpha j}=\left\{\begin{aligned}
	e^{-\frac{||\mathbf{X}_{\alpha,i}-\mathbf{X}_{\alpha,j}||^2_2}{\sigma^2}}, & \quad \mbox{dis}_1(\mathbf{X}_{\alpha,i},\mathbf{X}_{\alpha,j})<p,\\
	 &\quad \mbox{dis}_2(p(\alpha,i),p(\alpha, j))<q;\\
	0, & \quad \mbox{otherwise.}
	\end{aligned}
	\right.
	\end{align}
	where $\sigma$, $p$ and $q$ are design parameters and $p(\alpha,i)$ is the position of the superpixel $i$ in layer $\alpha$.
\end{itemize}
Beyond the traditional Gaussian distance \cite{c47}, our intralayer connections
consider two conditions for determining the presence of links: 1) features between two nodes should be similar; and 2) two connected superpixels should be in a localized region in the HSI. 
The first condition ensures the similarity of connected nodes while the second condition emphasizes geometric closeness in the HSI. 
For an initial setup of the parameters, we define $\mbox{dis}_1$ using
$\ell_2$-norm, and define $\mbox{dis}_2$ as the Euclidean distance between
the respective centroids of two superpixels. In terms of design parameters, 
we set $p$ as the mean of all pairwise similarities and tune the parameters $q$, $\sigma$ based on the specific dataset.

\subsection{MLN-based Spectral Clustering for HSI Segmentation}
Spectral clustering is an efficient method for unsupervised HSI segmentation \cite{c22}. Modeling HSI by a normal graph before spectral clustering, significant improvement is possible owing to its power in capturing the underlying structures \cite{c48,c49}. However, by representing HSI by a single-layer graph,
distinction of individual spectrum bands might be overlooked. To capture the heterogeneous spectral-spatial structure in HSI, we propose to segment the HSI based on the M-GSP spectral clustering as follows.

Consider an HSI $\mathbf{X}\in\mathbb{R}^{K}\times\mathbb{R}^{N}$ with $K$ spectrum frames and $N$ superpixels. To implement the M-GSP spectral clustering, we first construct an $M$-layer network the with adjacency tensor $\mathbf{A}\in \mathbb{R}^M\times \mathbb{R}^N\times \mathbb{R}^M \times \mathbb{R}^N$ 
according to
Eqs.~(\ref{inter})-(\ref{intra}). 
We then apply  HOSVD to obtain the singular tensors $\mathbf{F}_s=[\mathbf{f}_1\cdots\mathbf{f}_M]\in\mathbb{R}^M\times 
\mathbb{R}^M$ and $\mathbf{E}_s=[\mathbf{e}_1\cdots\mathbf{e}_N]\in\mathbb{R}^N\times \mathbb{R}^N$ to characterize the bands and superpixels, respectively, according to Eq.~(\ref{decomposeS}). Since we aim to segment superpixels into meaningful clusters, we focus on the entity-wise spectrum $\mathbf{E}_s$. Arranging $\mathbf{e}_i$ in the descending order of its corresponding singular value $\sigma_i$, i.e.,
\begin{equation}\label{sing}
	\sigma_i=||\mathbf{S}_{i_2=i}||,
\end{equation}
where $\mathbf{S}_{i_2=i}\in\mathbb{R}^{M}\times \mathbb{R}^{1}\times \mathbb{R}^{M} \times \mathbb{R}^{N}$ is the subtensor of the core tensor  $\mathbf{S}$ in Eq. (\ref{decomposeS}) by freezing the second order $i_2=i$, we pick the first $P$ singular vectors to preserve the most critical information for HSI based on a largest
gap among the singular values. Clustering based on the $P$ selected singular vectors and labeling each pixel within the superpixel, we can obtain a  
segmentation of the given HSI. The major process of MLN-based unsupervised segmentation is provided in Algorithm \ref{ALO1}.

\begin{algorithm}[htb]
	\begin{algorithmic}[1] 
		\caption{MLN-based Unsupervised HSI Segmentation (MLN-SC)}\label{ALO1}
		\STATE {\bf{Input}}: HSI $\mathbf{I}\in\mathbb{R}^{K} \times\mathbb{R}^T$ with $K$ frames and $T$ pixels, and the number of clusters $Q$;
		\STATE Construct $N$ superpixels for HSI based on ERS algorithm as $\mathbf{X}\in\mathbb{R}^{K} \times\mathbb{R}^N$;
		\STATE Divide bands into $M$ clusters based on $k$-means clustering;
		\STATE Construct an $M$-layer network with adjacency tensor $\mathbf{A}\in\mathbb{R}^M\times \mathbb{R}^N\times \mathbb{R}^M \times \mathbb{R}^N$ as Eq. (\ref{inter}) and Eq. (\ref{intra});
		\STATE Implement HOSVD on $\mathbf{A}$ to obtain the entity-wise singular tensors $\mathbf{E}_s=[\mathbf{e}_1\cdots\mathbf{e}_N]\in\mathbb{R}^N\times \mathbb{R}^N$ in the descending order of the $2$-mode singular values $\sigma_i$ as Eq. (\ref{sing});
		\STATE Select the first $P$ singular tensors as $\mathbf{P}_k=[\mathbf{e}_1,\cdots,\mathbf{e}_P]\in\mathbb{R}^N\times \mathbb{R}^P $ based on the largest singular gap;
		\STATE Cluster the rows of $\mathbf{P}_k$ into $Q$ groups based on $k$-means clustering;
		\STATE Cluster the superpixel $i$ into group $j$ if the $i$-th row of $\mathbf{P}_k$ is in group $j$;
		\STATE Label the pixels as the same cluster of its superpixel;
		\STATE  {\bf{Output}}: Segmented HSI.
	\end{algorithmic}
\end{algorithm}
\begin{figure*}[t]
	\centering
	\includegraphics[width=6.5in]{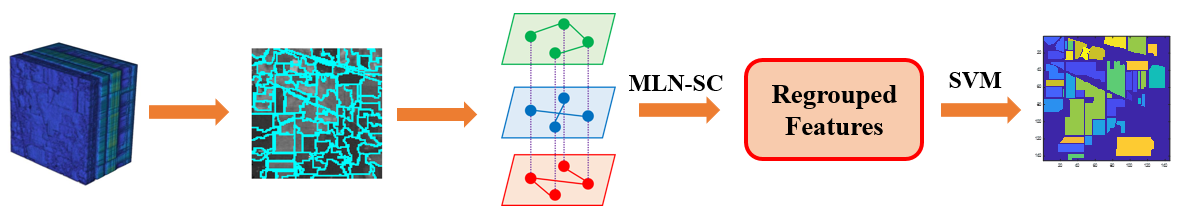}
	\caption{Scheme of Single-Resolution Segmentation.}
	\label{sig-ch}
\end{figure*}
\subsection{Discussion}
Before we dive further to develop 
MLN-based semi-supervised HSI segmentation, we provide a short conceptual
discussion on the M-GSP singular tensors. Within the context of GSP, how to capture  layer-wise and entity-wise information efficiently has become a significant topic
of research due to growing interests in spatial-temporal datasets. 
The authors of \cite{c48} have proposed a two-step graph Fourier transform (GFT) 
to process spatial-temporal graphs, by applying GFT in the spatial domain (intralayer) first and then in the temporal domain (interlayer). 
However, this approach does not consider the structural correlations 
between inter- and intra- layers. 
The authors of \cite{c49} developed a joint time-vertex Fourier transform (JFT) 
by implementing GFT and discrete Fourier transform (DFT) consecutively. 
However, the DFT block limits the structure of interlayer connections 
to a path graph. More recently, the authors of \cite{c50} proposed 
a tensor-based multi-way graph signal processing framework (MWGSP) 
on the product graph.  MWGSP constructs separate factor graphs for each mode of a tensor-represented signals and defines a joint spectrum that
combines spectra of all factor graphs. One limitation is that 
the combination process requires a homogeneous structure for each graph layer
and does not accommodate potentially different band-wise features in HSI. 

\begin{figure}[t]
	\centering
	\subfigure[Indian Pines]{
		\label{s1}
		\includegraphics[height=3.4cm]{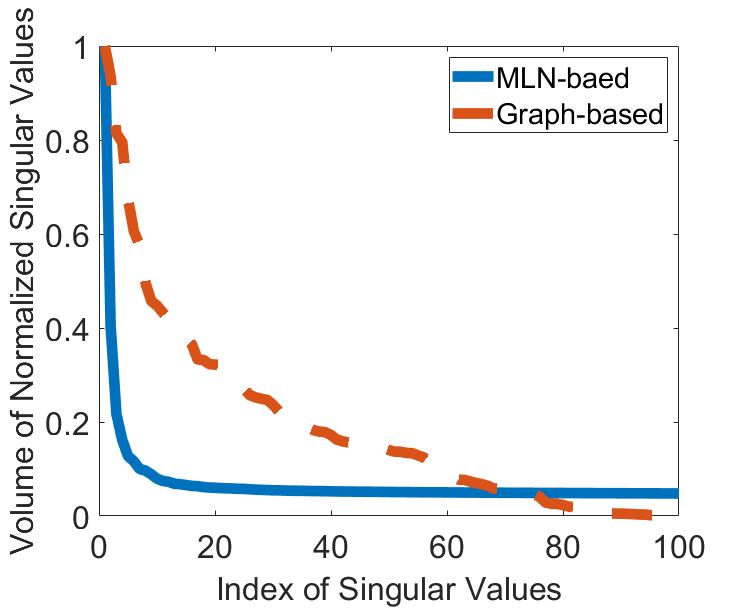}}
	\hfill
	\subfigure[Salinas]{
		\label{s2}
		\includegraphics[height=3.4cm]{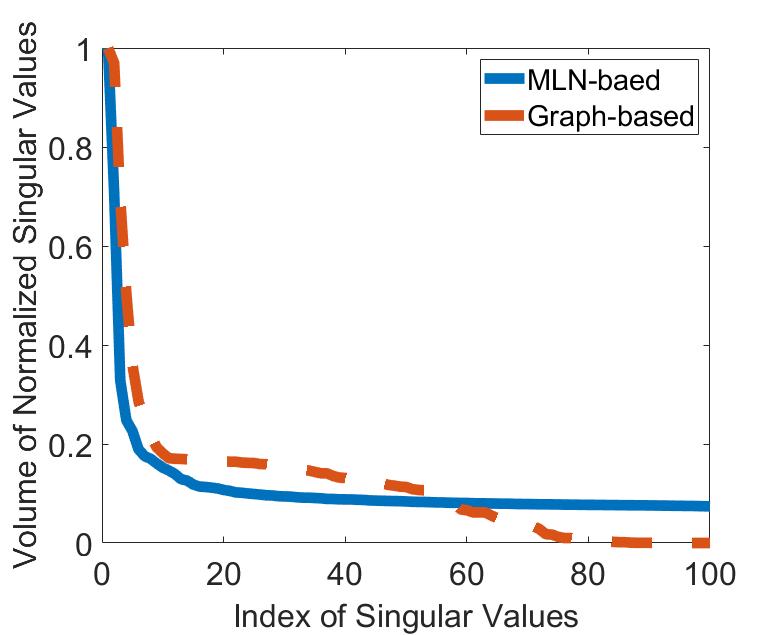}}
	\caption{Distribution of Entity-wise Singular Values.}
	\label{ex_s}
\end{figure}

By contrast, since we apply HOSVD in M-GSP for achieving
tensor-based representation, our M-GSP framework is
able to jointly process interlayer and intralayer connections of HSI.
Even when focusing on entity-wise singular tensors in the clustering process, 
we are still able to incorporate layer-wise information. To better understand the property of MLN-based singular tensors, we graphically illustrate the distribution of singular values compared to a graph-based model in Fig. \ref{ex_s}. 
As shown, the energy of MLN-based singular values are more concentrated 
in the first few dominant singular vectors in low frequency when compared against graph-based singular values.  This energy concentration indicates a more 
convenient and low degradation implementation of spectral clustering within our
proposed M-GSP framework.

\section{Semi-Supervised HSI Classification based on M-GSP Feature Extraction}\label{sup}

In this section, we introduce the semi-supervised HSI segmentation based on
M-GSP feature extraction. 
\subsection{Single-Resolution of MLN-based HSI Segmentation}
We start with the single-resolution of the superpixels. In the superpixel-based classification, superpixel resolutions affect the final performance: finer resolution could capture more details whereas coarse resolution captures the global information more efficiently. To benefit from both fine and coarse resolutions, 
we introduce the MLN-based spectral clustering on the fine resolution 
to regroup superpixels into a coarse resolution (the number of regrouped superpixels should still be larger than the number of classes) and use the regrouped features as classifier inputs. Here, we apply SVM to classify the regrouped superpixels. The concept of our single-resolution HSI segmentation (MLN-SRC) is 
illustrated by Fig. \ref{sig-ch}, and the major steps are described in Algorithm \ref{ALO2}.

\begin{algorithm}[htb]
	\begin{algorithmic}[1] 
		\caption{Single-Resolution HSI Segmentation (MLN-SRC)}\label{ALO2}
		\STATE {\bf{Input}}: HSI $\mathbf{I}\in\mathbb{R}^{K} \times\mathbb{R}^T$ with $K$ frames and $T$ pixels;
		\STATE Construct $N$ superpixels for HSI based on ERS algorithm as $\mathbf{X}\in\mathbb{R}^{K} \times\mathbb{R}^N$;
		\STATE Divide bands into $M$ clusters based on $k$-means clustering;
		\STATE Construct an $M$-layer network with adjacency tensor $\mathbf{A}\in\mathbb{R}^M\times \mathbb{R}^N\times \mathbb{R}^M \times \mathbb{R}^N$ as Eq. (\ref{inter}) and Eq. (\ref{intra});
		\STATE Implement MLN-based spectral clustering to regroup the superpixels into $D$ clusters, and combine the features of pixels within the same cluster as the regrouped features, i.e.,
		$\mathbf{X}_R\in\mathbb{R}^{K} \times\mathbb{R}^D$;
		\STATE Input $\mathbf{X}_R$ into SVM for the classification of regrouped superpixels;
		\STATE Label the pixels as the same class of its superpixel;
		\STATE  {\bf{Output}}: Segmented HSI.
	\end{algorithmic}
\end{algorithm}

The benefits of the proposed MLN-SRC include:
\begin{itemize}
	\item Against a singe resolution of coarse superpixels, the MLN-SRC implements an analysis step over a fine resolution and is capable of capturing detailed features. 
	Against a single resolution of fine superpixels, the MLN-SRC substantially reduces pixel number and enhances robustness of 
	the feature inputs to the classifier. 
 		Too many superpixels may 
	make the features less distinctive and over-segment the regions,
	whereas too few superpixels may lead to boundary ambiguity.
	
	\item Traditional graph-based superpixel segmentation only captures
	a single-layer structure. MLN-GSP regrouping 
	could reveal additional feature information across the heterogeneous 
	multi-band structures.
	\item In traditional superpixel segmentation, the distinct regions are usually labeled as different superpixels.
	However, in MLN-SRC, superpixels from
	different regions may have the same labels depending on
	clustering results. Thus, regrouped features can involve 
	similar pixels that cover a large distance and potentially
generate more features.
	\item  MLN-SRC can be easily integrated with other feature extraction or selection algorithms. Dimension reduction techniques such as
 PCA and ICA can potentially improve the performance when applied on
features and feature groups generated by MLN-SRC. 
\end{itemize}

\begin{figure}[t]
	\centering
	\subfigure[Indian Pines]{
		\label{s11}
		\includegraphics[height=4cm]{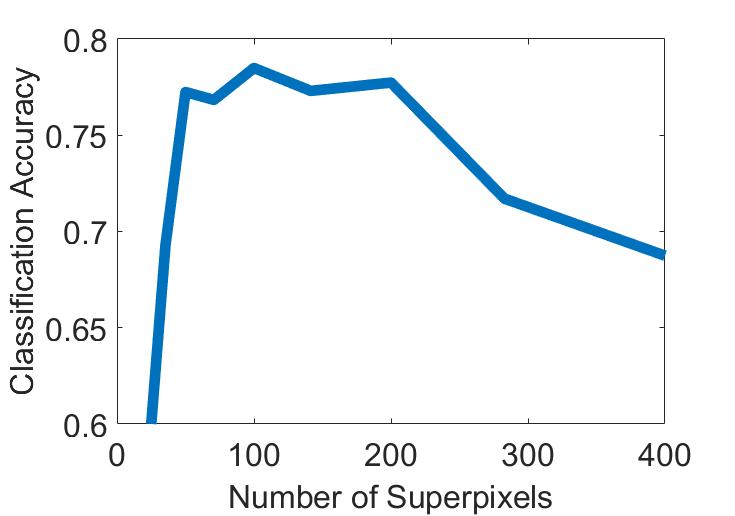}}
	\hspace{4cm}
	\subfigure[Salinas]{
		\label{s21}
		\includegraphics[height=4cm]{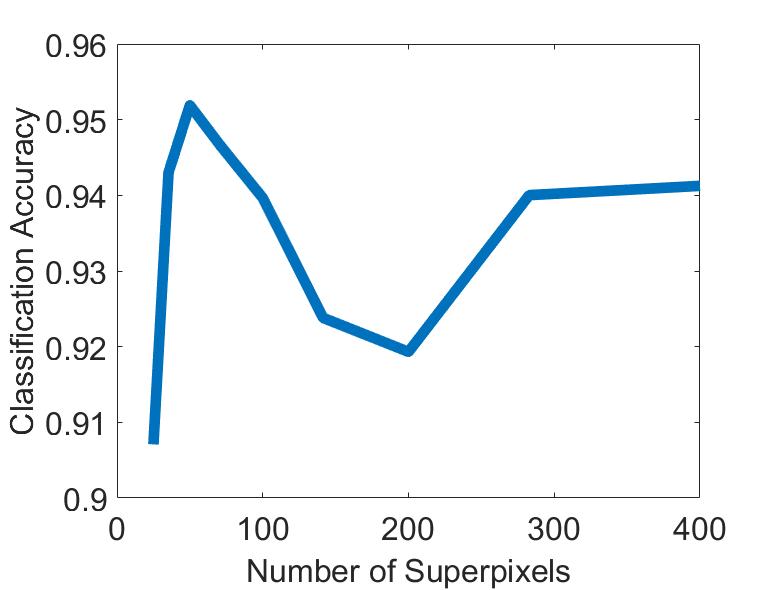}}
	\caption{Accuracy of MLN-SRC over Different Initial Resolutions.}
	\label{ex_s1}
\end{figure}
\subsection{Multi-Resolution of MLN-based HSI Segmentation}
\subsubsection{Multi-Resolution Structure}
Although MLN-based spectral clustering can regroup small superpixels into larger ones and to benefit from both fine and coarse resolutions in MLN-SRC, the initial resolution setting of superpixels still affects final performance. As Fig. \ref{ex_s1} shows, different initial resolutions can lead to different 
levels of accuracy. 
It is practically difficult 
to determine the optimal initial number of superpixels.

Similar to \cite{c15}, we consider a multi-resolution structure of classification (MLN-MRC) shown as Fig. \ref{supervised}. In this framework, we examine several different initial resolutions of superpixels. MLN-SRC is applied to
each initial resolution to regroup the superpixels with
a same reduction ratio in group numbers, i.e., $70\%$ of
initial superpixels. Applying
SVM to classifying the multiple regrouped superpixels, we fuse the results from different initial resolutions in final segmentation. 
The algorithm is described in Algorithm \ref{ALO3}.

\begin{algorithm}[t]
	\begin{algorithmic}[1] 
		\caption{Multi-Resolution HSI Segmentation (MLN-MRC)}\label{ALO3}
		\STATE {\bf{Input}}: HSI $\mathbf{I}\in\mathbb{R}^{K} \times\mathbb{R}^T$ with $K$ frames and $T$ pixels;
		\STATE Construct multiple resolutions of superpixels for HSI;
		\STATE Construct different multilayer networks for each resolution of HSI;
		\STATE Implement MLN-SRC for each resolution to obtain the sub-results;
		\STATE Fuse all the sub-results for the final segmentation;
		\STATE  {\bf{Output}}: Segmented HSI.
	\end{algorithmic}
\end{algorithm}
Although the multi-resolution structures have been considered in literature, 
MLN-MRC exhibits two major distinctions. First, we apply a novel MLN-based clustering algorithm to regroup the superpixels and generate new features for classification. Second, we provide several novel decision fusion strategies, based on both confidence score and graph structures to be discussed below. 

\subsubsection{Decision Fusion}
In \cite{c15}, a decision fusion scheme based on majority voting (MV) is proposed. In this method, the label $l$ of a specific pixel is determined by
\begin{equation}
	l={\arg\max}_i \sum_{j=1}^{C} w_j\cdot \delta(l_j),\label{eq:fusing}
\end{equation}
where $C$ is the number of distinct resolutions, $l_j$ is the label of the pixel in resolution $j$, $w_j$ is the
voting strength, and $\delta(l_j)=1$ if $l_j=i$; otherwise, $\delta(l_j)=0$.
Note that a basic majority voting based on equal strength $w_j={C}^{-1}$
applies the same strength to different resolutions but would ignore the 
difference of multiple resolutions. To improve decision fusion, 
we introduce several novel strategies for the decision fusion.

\begin{table*}[t]
	\centering
	\caption{Statistics of Different HSI Datasets}
	\begin{tabular}{|l|l|l|l|ll|}
		\hline
		HSI      & Pixel Size & \# of Spectrum Bands & \# of Classes & \multicolumn{2}{l|}{\# of Labeled Samples} \\ \hline
		IndianP  & 145$\times$145    & 200                  & 16            & \multicolumn{2}{l|}{10249}                 \\ \hline
		PaviaU   & 610$\times$340    & 103                  & 9             & \multicolumn{2}{l|}{42776}                 \\ \hline
		Salinas  & 512$\times$217    & 204                  & 16            & \multicolumn{2}{l|}{54129}                 \\ \hline
		SalinasA & 83$\times$86      & 204                  & 6             & \multicolumn{2}{l|}{5348}                  \\ \hline
	\end{tabular}
\label{hsi_sta}
\end{table*}

\begin{figure*}[t]
	\centering
	\subfigure[Indian Pines]{
		\label{g1}
		\includegraphics[height=3cm]{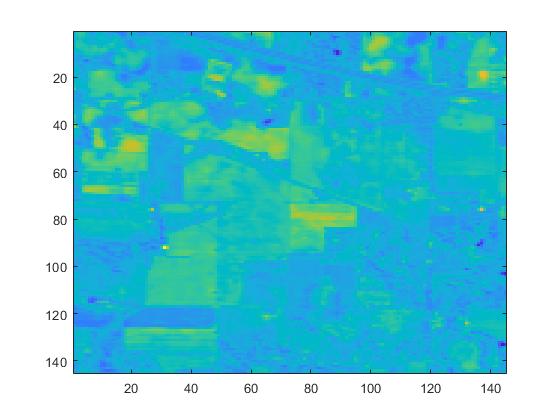}}
	\subfigure[Pavia University]{
		\label{g2}
		\includegraphics[height=3cm]{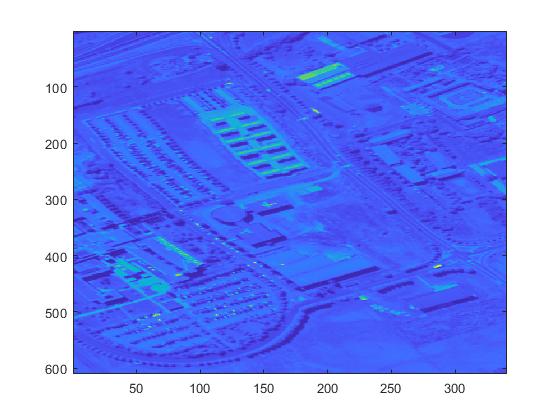}}
	\subfigure[Salinas]{
		\label{g3}
		\includegraphics[height=3cm]{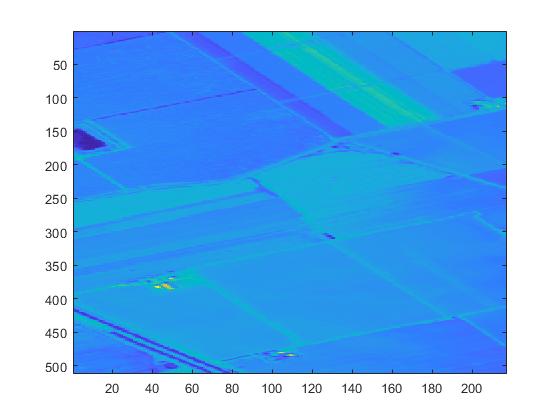}}
	\subfigure[SalinasA]{
		\label{g4}
		\includegraphics[height=3cm]{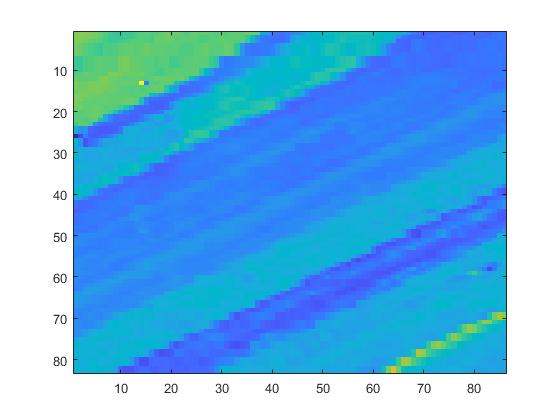}}
	\caption{Mean of HSI over Spectrum Dimension.}
		\label{mn}
\end{figure*}

\begin{itemize}
	\item \textit{Validation Accuracy} (VA): One intuitive way to design decision strength is based on validation accuracy. Here, we can apply the validation accuracy directly as the weighting strength $w_j$ for resolution $j$
	to fuse the decision according to Eq.~(\ref{eq:fusing}).
	
	\item \textit{Decision Value} (DV): As one alternative, the decision probability for each class of pixels can be used as the weight. In multi-class SVM, the predicted label is determined according to the decision value $\mathbf{p}\in\mathbb{R}^{C}$, where $C$ is the number of classes \cite{c53}. Let $\mathbf{p}_{ij}$ be the decision value of pixel $i$ in $j$th resolution. 
We set the weight of $l_{ij}$ to 
	 \begin{equation}
	 	w_{ij}=\max_k \mathbf{p}_{ij}(k).
	 \end{equation}
Unlike validation accuracy which is the same for all pixels in each
resolution, this weight based on decision value may vary even for pixels
at the same resolution.

	\item \textit{Graph Total Variation} (TV): Graph-based metrics can serve 
	as weights. For a robust setup of superpixels, signals should be smooth
	and exhibit stable underlying graph structure. 
	To this end, we introduce 
	graph-based total variation to measure smoothness. Given a superpixel segmentation $j$ of a HSI with $N$ superpixels and $K$ spectrum frames, we regenerate the features of each superpixel by averaging all pixels within. 
We then construct a single-layer graph based on Gaussian distance to measure similarity between different superpixels. Defining a Laplacian matrix by $\mathbf{L}=\mathbf{D-A}$ where $\mathbf{D}$ is the degree matrix and $\mathbf{A}$ is the adjacency matrix, the total variation of the feature signal $\mathbf{s}_p\in\mathbb{R}^{N}$ for the $p$th band frame over $\mathbf{L}$ is 
	\begin{equation}\label{tv}
		\mbox{TV}_p=||\mathbf{s}_p-\frac{1}{|\lambda_{\max}|}\mathbf{Ls}_p||_2^2,
	\end{equation}
where $\lambda_{\max}$ is the largest eigenvalue of $\mathbf{L}$.
Total variation describes the propagation differences between two steps. A smaller total variation indicates a more smooth signal. 
With $K$ frames in total, final smoothness for resolution $j$ is defined as $\mbox{SM}_j=\frac{1}{K}\sum_p \mbox{TV}_p$.
Since we prefer a larger weight for the smooth signal, the final weight of the resolution $j$ is defined as
	\begin{equation}
		w_j=e^{-SM_j}.
	\end{equation}

	\item \textit{Von Neumann Entropy} (VN): 
The stability of the underlying graph structure can also indicate the confidence level of a specific superpixel resolution. 
In quantum theory \cite{c54}, a pure state leads to a zero Von Neumann entropy. The entropy is larger if there are more mixed states in the system. Similarly, in our HSI analysis, since we prefer a stable system or a pure state on the underlying graph, the weight should be larger if the Von Neumann entropy is smaller. 
Consider the Von Neumann entropy introduced to evaluate the graph stability \cite{c54}. Similar to total variation, a Laplacian matrix $\mathbf{L}$ can be defined with adjacency matrix $\mathbf{A}=(a_{pq})$ for the $j$th resolution. 
First, define $c={1}/({\sum_{p,q}a_{pq}})$ and rescale the Laplacian matrix 
	\begin{equation}
		\mathbf{L}_G=c\cdot (\mathbf{D-A}),
	\end{equation}
We can define the weight for the $j$th resolution as 
	\begin{equation}
	w_j=e^{-h_j}.
	\end{equation}
based on the Von Neumann entropy
	\begin{equation}
		h_j=-Tr[\mathbf{L}_G\log_2\mathbf{L}_G].
	\end{equation}

\end{itemize}

Note that, here we provided several possible alternatives for the weights of decision fusion.  The performances of the various proposed fusion strategies will be presented in Section \ref{exp-2}. We plan to investigate more M-GSP based approaches to result
fusion in future works.

\section{Experimental Results}\label{exp}
We now test the performance of the proposed unsupervised and semi-supervised segmentation approaches in several well-known datasets to
demonstrate the efficacy of M-GSP in HSI analysis. We also comparatively test
the performance of various different fusion decisions. 

\subsection{Dataset}\label{dat}

We test the performances of the proposed methods based on
four public HSI datasets accessible from website \footnote{ \url{http://www.ehu.eus/ccwintco/index.php/Hyperspectral_Remote_Sensing_Scenes}}.
The first HSI is \textit{Indian Pines} (IndianP) scene 
originally gathered by AVIRIS sensors over an agricultural field. The second 
HSI dataset is the \textit{University of Pavia} (PaviaU) acquired by ROSIS sensor. Note that, some of the samples in PaviaU contain no information and have to be discarded before analysis. Two other HSIs used in the experiments are the \textit{Salinas Scene} (Salinas) and \textit{Salinas-A Scene} (SalinasA) datasets, which were collected by the 224-band AVIRIS sensor over Salinas Valley, California, and exhibit high spatial resolution. For each dataset, we have groundtruth 
classes for part of samples. 

\begin{figure*}[t]
	\centering
	\subfigure[Indian Pines]{
		\label{g11}
		\includegraphics[height=3cm]{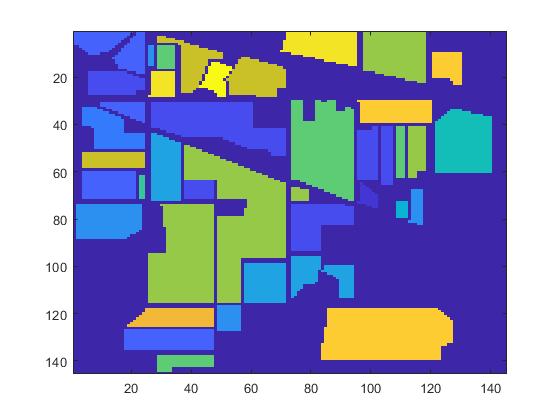}}
	\subfigure[Pavia University]{
		\label{g21}
		\includegraphics[height=3cm]{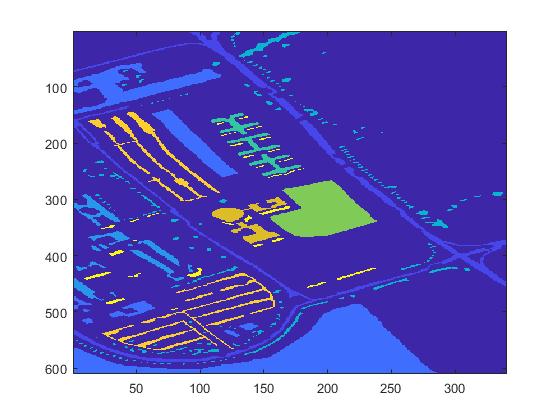}}
	\subfigure[Salinas]{
		\label{g31}
		\includegraphics[height=3cm]{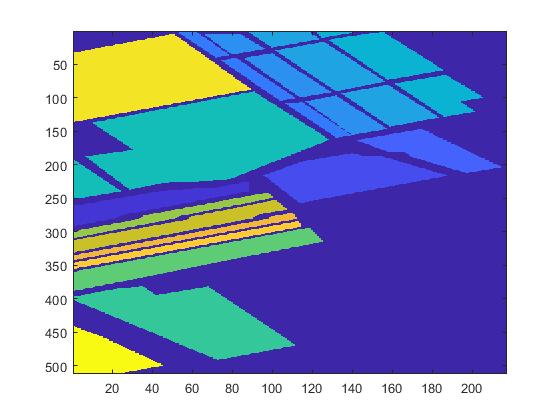}}
	\subfigure[SalinasA]{
		\label{g41}
		\includegraphics[height=3cm]{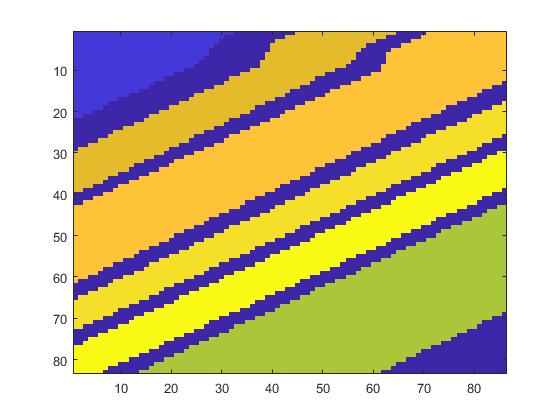}}
	\caption{Ground Truth of Class Labels}
	\label{gt}
	\vspace{-4mm}
\end{figure*}

For these HSIs, Table~\ref{hsi_sta} provides vital statistics 
and we provide visual illustration of the geometric plots in Fig.~\ref{mn} and 
Fig.~\ref{gt}. Note that, Fig.~\ref{gt} treats the unlabeled groundtruth
samples
as backgrounds with the same class label.
Interested readers can find more information on the HSI datasets 
at the website$^1$.

\begin{figure}[t]
	\centering
	\subfigure[K-means]{
		\label{unn11}
		\includegraphics[height=2cm]{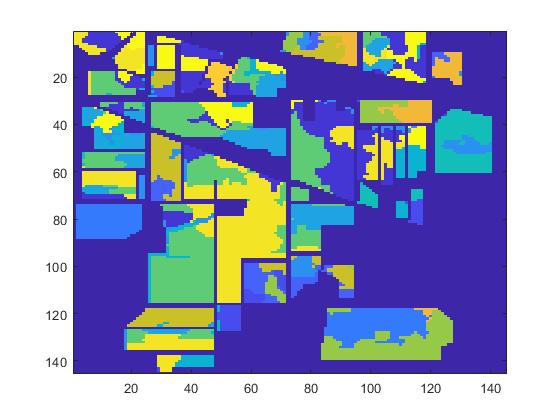}}
	\subfigure[GSP]{
		\label{unn12}
		\includegraphics[height=2cm]{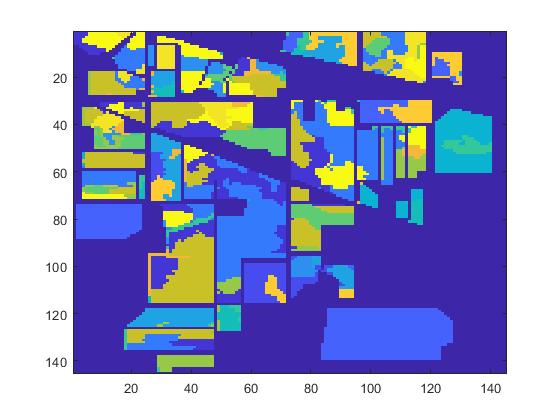}}
	\subfigure[M-GSP]{
		\label{unn13}
		\includegraphics[height=2cm]{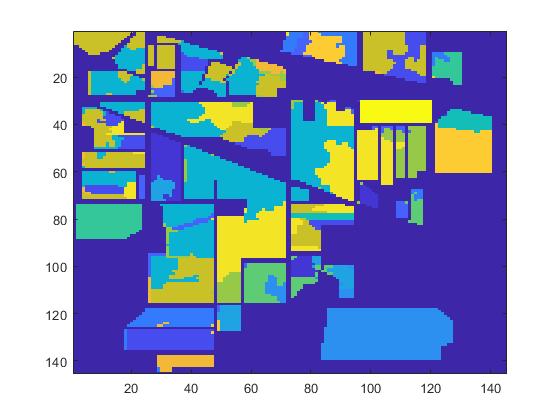}}
	\caption{Segmented Results of Indian Pines.}
	\label{unn1}
\end{figure}

\begin{figure}[t]
	\centering
	\subfigure[K-means]{
		\label{unn21}
		\includegraphics[height=2cm]{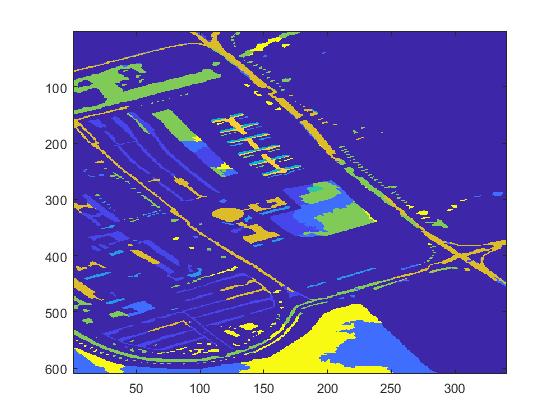}}
	\subfigure[GSP]{
		\label{unn22}
		\includegraphics[height=2cm]{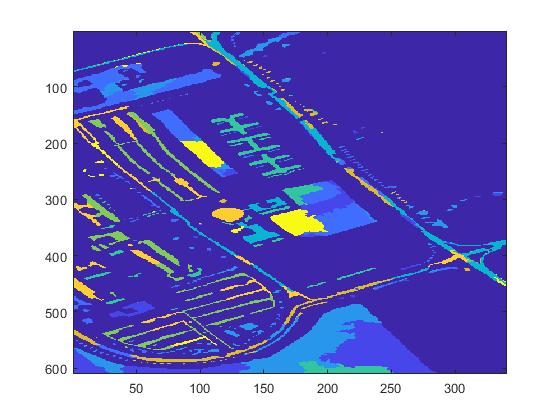}}
	\subfigure[M-GSP]{
		\label{unn23}
		\includegraphics[height=2cm]{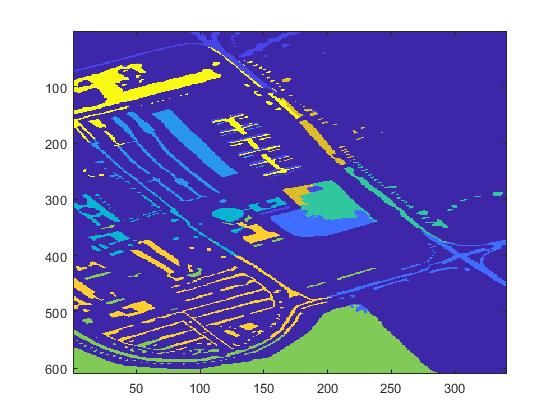}}
	\caption{Segmented Results of Pavia University.}
	\label{unn2}
\end{figure}
\begin{figure}[t]
	\centering
	\subfigure[K-means]{
		\label{unn31}
		\includegraphics[height=2cm]{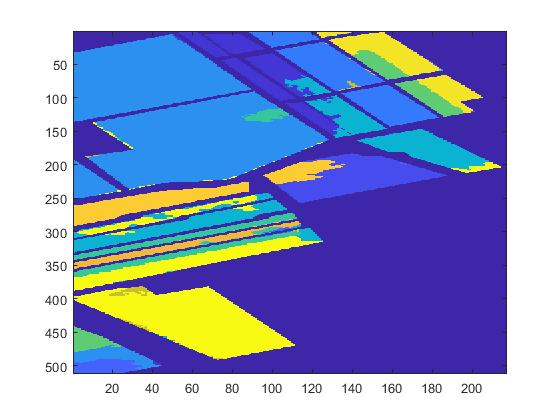}}
	\subfigure[GSP]{
		\label{unn32}
		\includegraphics[height=2cm]{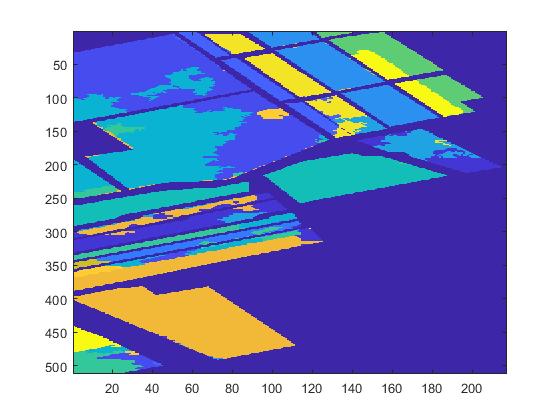}}
	\subfigure[M-GSP]{
		\label{unn33}
		\includegraphics[height=2cm]{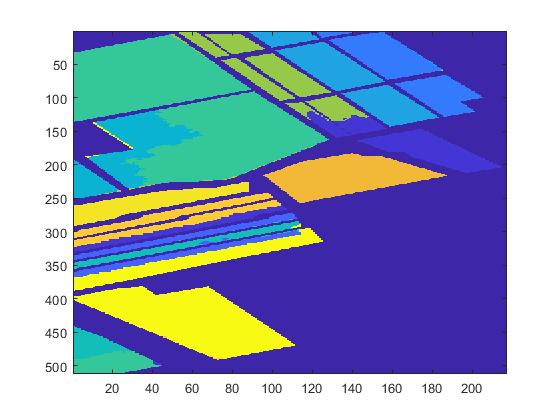}}
	\caption{Segmented Results of Salinas.}
	\label{unnn3}
\end{figure}

\begin{figure}[t]
	\centering
	\subfigure[K-means]{
		\label{unn41}
		\includegraphics[height=2cm]{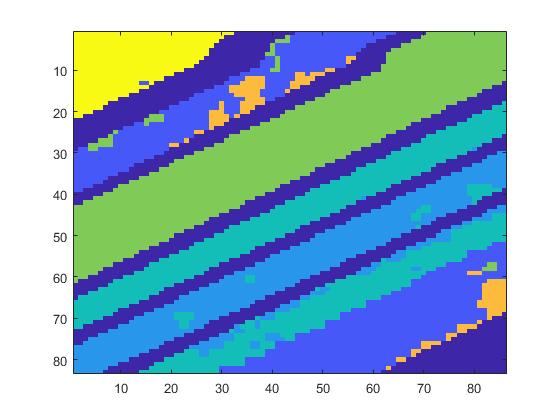}}
	\subfigure[GSP]{
		\label{unn42}
		\includegraphics[height=2cm]{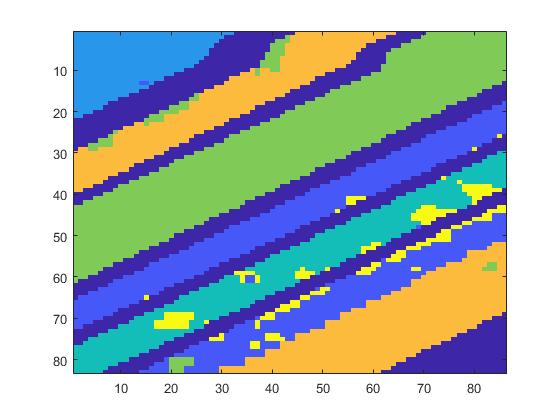}}
	\subfigure[M-GSP]{
		\label{unn43}
		\includegraphics[height=2cm]{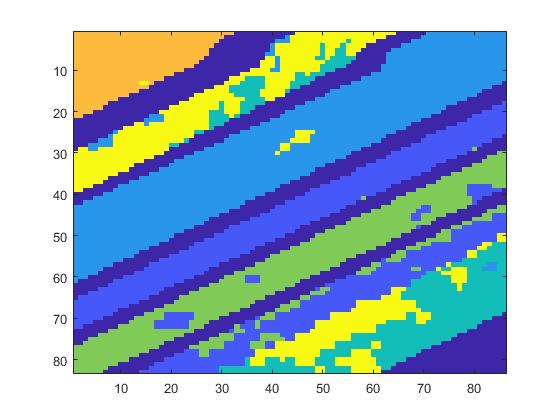}}
	\caption{Segmented Results of Salinas A.}
		\label{unn4}
\end{figure}

\subsection{Unsupervised HSI Segmentation}
In this part, we first test the performance of unsupervised HSI segmentation. 
Although our proposed MLN-based method is amenable to various 
sophisticated clustering approaches, we find it easier to demonstrate
it by using a basic spectral clustering scheme. In this section, 
we mainly use comparison of basic methods, such as $k$-means clustering and GSP-based spectral clustering \cite{c35}, to demonstrate the power of M-GSP in processing
the HSI datasets. 
We will illustrate the benefits of multilayer network models and M-GSP 
in comparison with other more advanced methods in semi-supervised segmentation 
in Section \ref{exp-2}.

To validate the performance of different methods, we consider two 
experimental setups. In the first scenario, we carry out
unsupervised segmentation on all data samples and evaluate the 
overall visualization results of labeled samples. In the second stage, we process
all data samples but focus on detecting boundaries (edges) of each cluster,
in terms of both visualization results and numerical accuracy.

For fair comparison, we segment the HSI into $N$ superpixels first before 
applying respective clustering algorithms thereupon. For the GSP-based method, we construct the graph $\mathbf{W}\in\mathbb{R}^{N\times N}$ using Gaussian distance
	\begin{align}
W_{i j}=\left\{\begin{aligned}
e^{-\frac{||\mathbf{s}_{i}-\mathbf{s}_{j}||^2_2}{\sigma^2}}, & \quad ||\mathbf{s}_{i}-\mathbf{s}_{j}||^2_2\leq \tau,\\
0, & \quad otherwise,
\end{aligned}
\right.
\end{align}
where $\mathbf{s}_i$ denotes the feature of $i$th superpixel. 
The threshold $\tau$ is set to the statistical 
mean of all pairwise distances among superpixels, and $\sigma$ 
is tunable according to specific datasets. 
For M-GSP based methods, we construct the multilayer network with $M=10$ layers, and calculate the distance based on Eq.~(\ref{inter}) and Eq.~(\ref{intra}). The parameter $p$ in Eq. (\ref{intra}) is also set to the statistical mean of all pairwise intralayer feature distances, and $q=100$ is used. 
We select the number of key spectra based on the largest gap of singular values. 
We summarize the results below. 
\begin{figure*}[t]
	\centering
	\subfigure[Ground Truth]{
		\label{e11}
		\includegraphics[height=3cm]{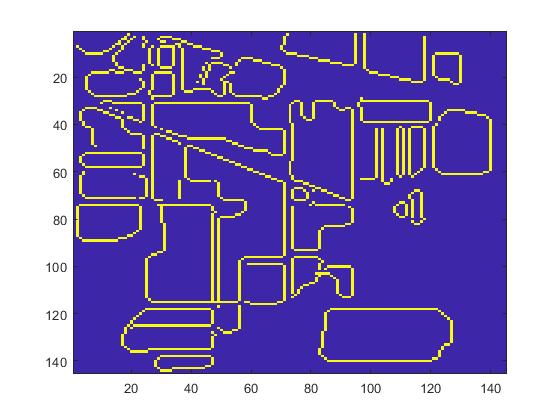}}
	\subfigure[K-means]{
		\label{e12}
		\includegraphics[height=3cm]{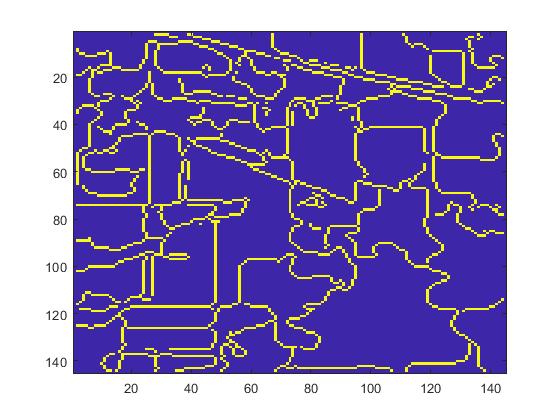}}
	\subfigure[GSP]{
		\label{e13}
		\includegraphics[height=3cm]{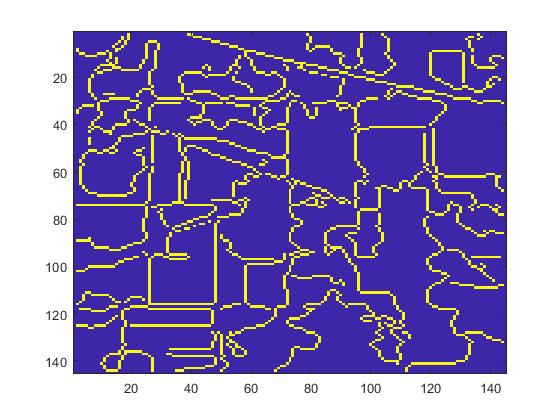}}
	\subfigure[M-GSP]{
		\label{e14}
		\includegraphics[height=3cm]{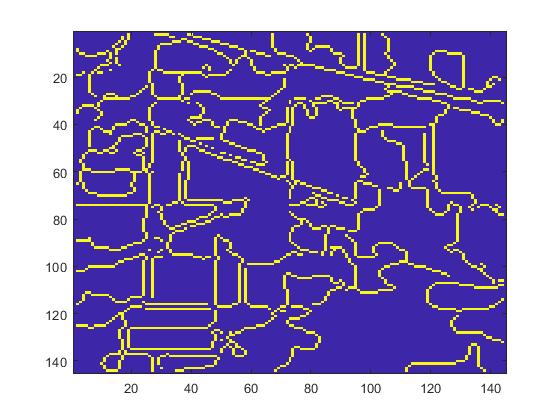}}
	\caption{Boundaries in Indian Pines.}
	\label{e1}
\end{figure*}

\begin{figure*}[t]
	\centering
	\subfigure[Ground Truth]{
		\label{e21}
		\includegraphics[height=3cm]{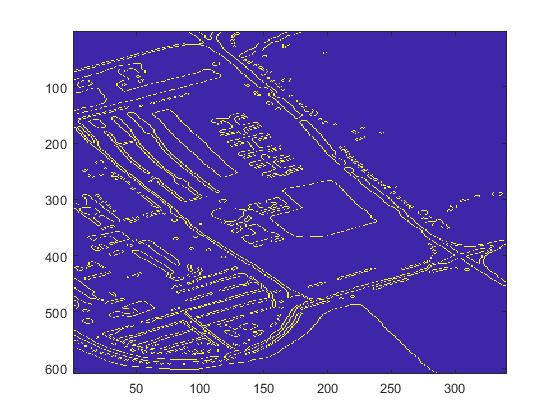}}
	\subfigure[K-means]{
		\label{e22}
		\includegraphics[height=3cm]{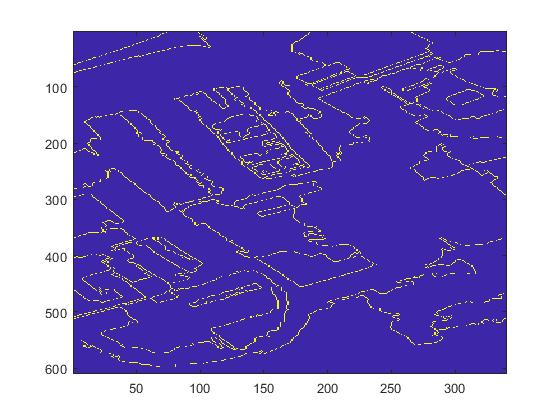}}
	\subfigure[GSP]{
		\label{e23}
		\includegraphics[height=3cm]{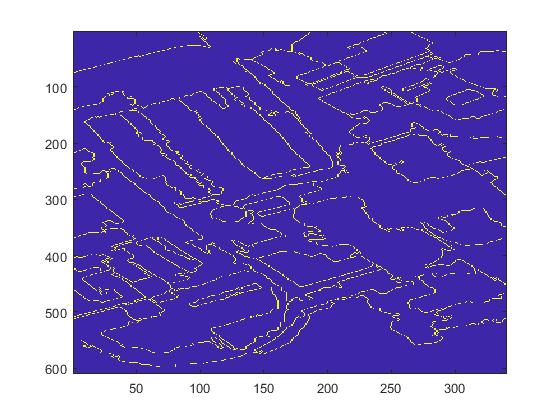}}
	\subfigure[M-GSP]{
		\label{e24}
		\includegraphics[height=3cm]{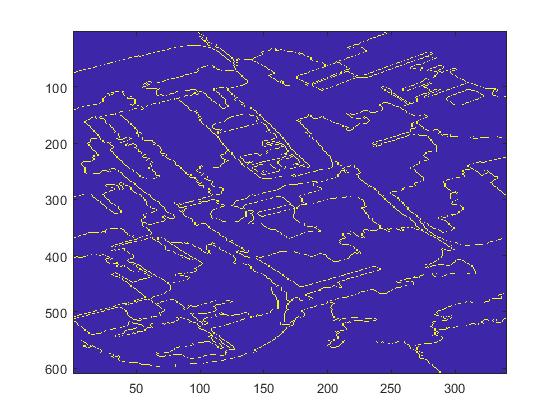}}
	\caption{Boundaries in Pavia University.}
	\label{e2}
\end{figure*}
\begin{figure*}[t]
	\centering
	\subfigure[Ground Truth]{
		\label{e31}
		\includegraphics[height=3cm]{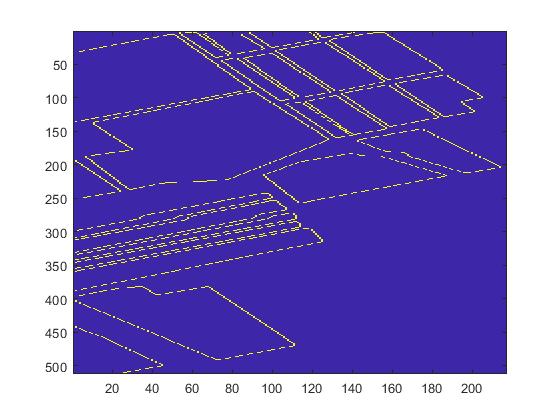}}
	\subfigure[K-means]{
		\label{e32}
		\includegraphics[height=3cm]{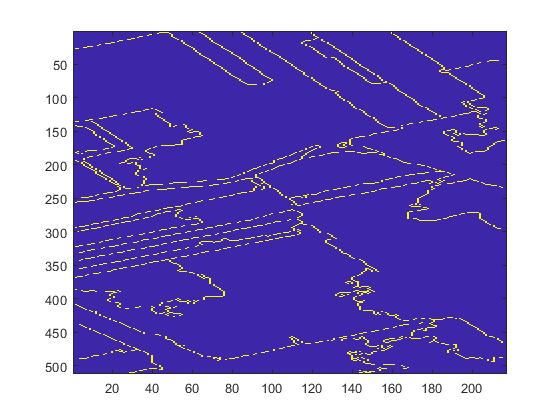}}
	\subfigure[GSP]{
		\label{e33}
		\includegraphics[height=3cm]{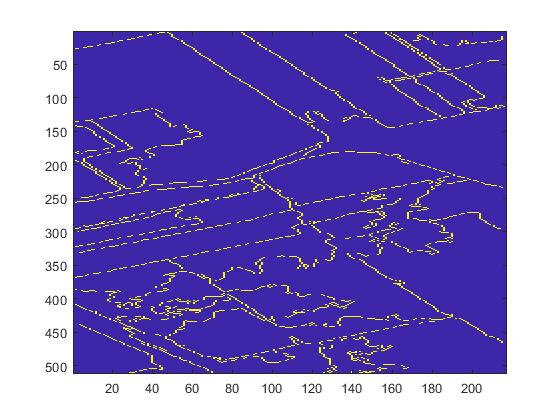}}
	\subfigure[M-GSP]{
		\label{e34}
		\includegraphics[height=3cm]{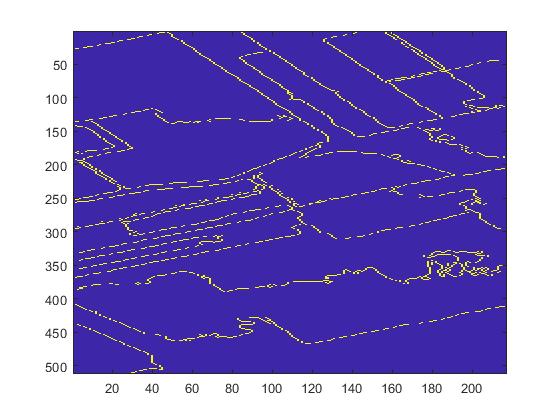}}
	\caption{Boundaries in Salinas.}
	\label{e3}
\end{figure*}

\subsubsection{Visualization of HSI Segmentation}
By setting $N=500$ for all tested HSIs,
Fig. \ref{unn1} - Fig. \ref{unn4} present the
visualization results of the three HSI segmentations using different
algorithm in comparison to the groundtruth data. 
These results show that, in general, the MLN-based spectral clustering 
(M-GSP) displays more stable segmentations than the
single layer GSP-based spectral clustering 
(GSP) as well as the $k$-means algorithm.   
Since it is harder to evaluate the details for too many classes, we 
can focus more on the exemplary \textit{Salinas A} dataset. 
In particular, Fig. \ref{g41} shows six different groundtruch classes. 
However, both $k$-means and GSP-based method failed to detect the class No.6, marked
by orange in Fig. \ref{unn41} and marked by yellow in Fig. \ref{unn42}, respectively. However, in Fig. \ref{unn43}, the MLN-GSP method successfully 
identified all six classes and delivered results that are closer to groundtruth. Recall that, from Fig. \ref{ex_s}, the MLN-based singular values are more 
concentrated, which provides the benefits of robustness in
spectral clustering.  Results from ``Indian Pines'' and ``Pavia University'' 
also show similarly stronger performance for M-GSP.  
These results collectively demonstrate improved efficiency of M-GSP in 
unsupervised HSI segmentation.

\subsubsection{Boundary Segmentation}\label{unsup2}
Although unsupervised methods may generate
meaningful segmentation different from groundtruth, we are still interested in how far the segmented results are from the true labels. Since it is inefficient to match all the clusters to the corresponding true labels, we focus 
on the boundaries of the segmentation. In the boundary detection, we set $N=100$ and define the accuracy as 
\begin{equation}\label{accc}
	Acc=\sum_{i=1}^T 1(L_i=\hat{L}_i)/T,
\end{equation}
\begin{table}[t]
	\caption{Accuracy of Segmentation Boundaries}
	\centering
	\begin{tabular}{|l|l|l|l|}
		\hline
		Data    & K-means & GSP    & M-GSP  \\ \hline
		IndianP & 0.8257  & 0.8298 & \textbf{0.8441} \\ \hline
		Salinas & 0.9208  & 0.9285 & \textbf{0.9409} \\ \hline
		PaviaU  & 0.9070  & 0.9088 & \textbf{0.9255} \\ \hline
	\end{tabular}
\label{accs}
\end{table}
where $T$ is the number of pixels in the HSI, $L_i$ is the true labels of edges for the $i$th pixel, $\hat{L}_i$ is the estimated labels, and  $1(\cdot)$ denotes the indicator function. The boundary results are visualized in Fig. \ref{e1} - Fig. \ref{e3}. Since we consider the unlabeled samples in the clustering process, there can be more details for the HSIs than their corresponding
groundtruth, especially for \textit{Pavia University} and \textit{Indian Pines}. However, we can still derive some benefits of M-GSP in boundary detection. As Fig. \ref{e3} shows, M-GSP can generate clearer edges, obviously on the top half image, while GSP and K-means appear to over-segment. 
We also present the accuracy defined in Eq.~(\ref{accc}) as Table~\ref{accs}. 
These results show that
M-GSP performs better than K-means and GSP. They demonstrate
the efficiency of MLN models in HSI analysis. 

It is important to note that
\begin{itemize}
    \item We do not claim M-GSP to be the best approaches for all scenarios.
Without impractically requiring hindsight to fine-tune various parameters 
for each HSI dataset to generate ``the best results'',
M-GSP delivers consistently strong and stable segmentation results 
for various HSI datasets by relying on some basic guidelines for
selecting parameters. 
    \item We present M-GSP as an alternative approach to HSI segmentation. Its
scalability and practical tuning allow easier
integration with other graph-based algorithms and deep learning
neural networks.
\end{itemize}

\begin{table*}[htb]
	\centering
	\caption{Overall Accuracy of Different HSI Segmentation}
	\label{oa}
	\begin{tabular}{|
			>{\columncolor[HTML]{FFFFFF}}l |
			>{\columncolor[HTML]{FFFFFF}}l |
			>{\columncolor[HTML]{FFFFFF}}l |
			>{\columncolor[HTML]{FFFFFF}}l |
			>{\columncolor[HTML]{FFFFFF}}l |
			>{\columncolor[HTML]{FFFFFF}}l |
			>{\columncolor[HTML]{FFFFFF}}l |
			>{\columncolor[HTML]{FFFFFF}}l |
			>{\columncolor[HTML]{FFFFFF}}l |
			>{\columncolor[HTML]{FFFFFF}}l |
			>{\columncolor[HTML]{FFFFFF}}l |
			>{\columncolor[HTML]{FFFFFF}}l |
			>{\columncolor[HTML]{FFFFFF}}l |
			>{\columncolor[HTML]{FFFFFF}}l |}
		\hline
		{\color[HTML]{000000} Data}                                              & {\color[HTML]{000000} TS/C} & {\color[HTML]{000000} Raw}    & {\color[HTML]{000000} PCA}    & {\color[HTML]{000000} ICA}    & {\color[HTML]{000000} LPP}    & {\color[HTML]{000000} NPE}    & {\color[HTML]{000000} LPNPE}  & {\color[HTML]{000000} LDA}    & {\color[HTML]{000000} LFDA}   & {\color[HTML]{000000} SPCA}   & {\color[HTML]{000000} MSPCA}           & {\color[HTML]{000000} MLN-SRC} & {\color[HTML]{000000} MLN-MRC}         \\ \hline
		\cellcolor[HTML]{FFFFFF}{\color[HTML]{000000} }                          & {\color[HTML]{000000} 5}    & {\color[HTML]{000000} 0.4488} & {\color[HTML]{000000} 0.4637} & {\color[HTML]{000000} 0.4521} & {\color[HTML]{000000} 0.5358} & {\color[HTML]{000000} 0.5368} & {\color[HTML]{000000} 0.6725} & {\color[HTML]{000000} 0.5995} & {\color[HTML]{000000} 0.5962} & {\color[HTML]{000000} 0.7734} & {\color[HTML]{000000} 0.7868}          & {\color[HTML]{000000} 0.7432}  & {\color[HTML]{000000} \textbf{0.7936}} \\ \cline{2-14} 
		\cellcolor[HTML]{FFFFFF}{\color[HTML]{000000} }                          & {\color[HTML]{000000} 10}   & {\color[HTML]{000000} 0.5577} & {\color[HTML]{000000} 0.5572} & {\color[HTML]{000000} 0.5712} & {\color[HTML]{000000} 0.7041} & {\color[HTML]{000000} 0.7049} & {\color[HTML]{000000} 0.7645} & {\color[HTML]{000000} 0.6930} & {\color[HTML]{000000} 0.6491} & {\color[HTML]{000000} 0.8576} & {\color[HTML]{000000} 0.8712}          & {\color[HTML]{000000} 0.8604}  & {\color[HTML]{000000} \textbf{0.8773}} \\ \cline{2-14} 
		\multirow{-3}{*}{\cellcolor[HTML]{FFFFFF}{\color[HTML]{000000} IndianP}} & {\color[HTML]{000000} 20}   & {\color[HTML]{000000} 0.6381} & {\color[HTML]{000000} 0.6297} & {\color[HTML]{000000} 0.6441} & {\color[HTML]{000000} 0.8026} & {\color[HTML]{000000} 0.7987} & {\color[HTML]{000000} 0.8351} & {\color[HTML]{000000} 0.7656} & {\color[HTML]{000000} 0.7401} & {\color[HTML]{000000} 0.9390} & {\color[HTML]{000000} \textbf{0.9569}} & {\color[HTML]{000000} 0.9226}  & {\color[HTML]{000000} 0.9545}          \\ \hline
		\cellcolor[HTML]{FFFFFF}{\color[HTML]{000000} }                          & {\color[HTML]{000000} 5}    & {\color[HTML]{000000} 0.6459} & {\color[HTML]{000000} 0.6526} & {\color[HTML]{000000} 0.6658} & {\color[HTML]{000000} 0.7086} & {\color[HTML]{000000} 0.6835} & {\color[HTML]{000000} 0.7612} & {\color[HTML]{000000} 0.7243} & {\color[HTML]{000000} 0.7467} & {\color[HTML]{000000} 0.7439} & {\color[HTML]{000000} 0.7849}          & {\color[HTML]{000000} 0.7561}  & {\color[HTML]{000000} \textbf{0.8236}} \\ \cline{2-14} 
		\cellcolor[HTML]{FFFFFF}{\color[HTML]{000000} }                          & {\color[HTML]{000000} 10}   & {\color[HTML]{000000} 0.7022} & {\color[HTML]{000000} 0.7015} & {\color[HTML]{000000} 0.7139} & {\color[HTML]{000000} 0.8129} & {\color[HTML]{000000} 0.8063} & {\color[HTML]{000000} 0.8255} & {\color[HTML]{000000} 0.8124} & {\color[HTML]{000000} 0.7895} & {\color[HTML]{000000} 0.8342} & {\color[HTML]{000000} \textbf{0.9167}} & {\color[HTML]{000000} 0.8398}  & {\color[HTML]{000000} 0.8896}          \\ \cline{2-14} 
		\multirow{-3}{*}{\cellcolor[HTML]{FFFFFF}{\color[HTML]{000000} PaviaU}}  & {\color[HTML]{000000} 20}   & {\color[HTML]{000000} 0.7585} & {\color[HTML]{000000} 0.7591} & {\color[HTML]{000000} 0.7665} & {\color[HTML]{000000} 0.8600} & {\color[HTML]{000000} 0.8569} & {\color[HTML]{000000} 0.8856} & {\color[HTML]{000000} 0.8500} & {\color[HTML]{000000} 0.8698} & {\color[HTML]{000000} 0.8938} & {\color[HTML]{000000} \textbf{0.9537}} & {\color[HTML]{000000} 0.8697}  & {\color[HTML]{000000} 0.9432}          \\ \hline
		\cellcolor[HTML]{FFFFFF}{\color[HTML]{000000} }                          & {\color[HTML]{000000} 5}    & {\color[HTML]{000000} 0.8179} & {\color[HTML]{000000} 0.8187} & {\color[HTML]{000000} 0.8175} & {\color[HTML]{000000} 0.8523} & {\color[HTML]{000000} 0.8486} & {\color[HTML]{000000} 0.9209} & {\color[HTML]{000000} 0.8903} & {\color[HTML]{000000} 0.8883} & {\color[HTML]{000000} 0.9442} & {\color[HTML]{000000} 0.9500}          & {\color[HTML]{000000} 0.9499}  & {\color[HTML]{000000} \textbf{0.9588}} \\ \cline{2-14} 
		\cellcolor[HTML]{FFFFFF}{\color[HTML]{000000} }                          & {\color[HTML]{000000} 10}   & {\color[HTML]{000000} 0.8524} & {\color[HTML]{000000} 0.8528} & {\color[HTML]{000000} 0.8574} & {\color[HTML]{000000} 0.8860} & {\color[HTML]{000000} 0.8899} & {\color[HTML]{000000} 0.9452} & {\color[HTML]{000000} 0.9146} & {\color[HTML]{000000} 0.8277} & {\color[HTML]{000000} 0.9678} & {\color[HTML]{000000} 0.9815}          & {\color[HTML]{000000} 0.9614}  & {\color[HTML]{000000} \textbf{0.9863}} \\ \cline{2-14} 
		\multirow{-3}{*}{\cellcolor[HTML]{FFFFFF}{\color[HTML]{000000} Salinas}} & {\color[HTML]{000000} 20}   & {\color[HTML]{000000} 0.8785} & {\color[HTML]{000000} 0.8779} & {\color[HTML]{000000} 0.8808} & {\color[HTML]{000000} 0.9061} & {\color[HTML]{000000} 0.9069} & {\color[HTML]{000000} 0.9589} & {\color[HTML]{000000} 0.9372} & {\color[HTML]{000000} 0.9356} & {\color[HTML]{000000} 0.9837} & {\color[HTML]{000000} 0.9904}          & {\color[HTML]{000000} 0.9840}  & {\color[HTML]{000000} \textbf{0.9915}} \\ \hline
	\end{tabular}
\end{table*}

\begin{table*}[t]
	\centering
	\caption{Peformance Under the Same Sets of Superpixel Resolutions}
	\label{hsi}
	\begin{tabular}{|l|lll|lll|lll|}
		\hline
		Data    & \multicolumn{3}{l|}{IndianP}                                                                  & \multicolumn{3}{l|}{Salinas}                                                                  & \multicolumn{3}{l|}{PaviaU}                                                                   \\ \hline
		TS/C    & \multicolumn{1}{l|}{5}               & \multicolumn{1}{l|}{10}              & 20              & \multicolumn{1}{l|}{5}               & \multicolumn{1}{l|}{10}              & 20              & \multicolumn{1}{l|}{5}               & \multicolumn{1}{l|}{10}              & 20              \\ \hline
		MSPCA   & \multicolumn{1}{l|}{0.6663}          & \multicolumn{1}{l|}{0.7476}          & 0.9406          & \multicolumn{1}{l|}{0.8521}          & \multicolumn{1}{l|}{0.9662}          & \textbf{0.9934} & \multicolumn{1}{l|}{0.7078}          & \multicolumn{1}{l|}{0.8447}          & \textbf{0.9420} \\ \hline
		MLN-MRC & \multicolumn{1}{l|}{\textbf{0.7094}} & \multicolumn{1}{l|}{\textbf{0.8062}} & \textbf{0.9455} & \multicolumn{1}{l|}{\textbf{0.9365}} & \multicolumn{1}{l|}{\textbf{0.9738}} & 0.9931          & \multicolumn{1}{l|}{\textbf{0.8218}} & \multicolumn{1}{l|}{\textbf{0.8590}} & 0.9334          \\ \hline
	\end{tabular}
\end{table*}

\subsection{Semi-supervised HSI Segmentation}\label{exp-2}
We next test M-GSP in (semi)-supervised HSI segmentation.

\subsubsection{Overall Accuracy}\label{oac}
Applying M-GSP based spectral clustering as feature extraction of HSI, we compare the proposed algorithms with several well known
feature extraction algorithms, including PCA \cite{d1}, ICA \cite{c16}, LPP \cite{d2}, NPE \cite{d3}, LP-NPE \cite{d4}, LDA \cite{c14}, LFDA \cite{d5}, SPCA \cite{c15} and MSPCA \cite{c15}. Notably, MSPCA is also an algorithm integrating multiple resolutions of superpixels. 
For the proposed MLN-SRC and MLN-MRC, we regroup the superpixels to $70\%$ of the original superpixel number, before extracting features based on M-GSP. 
Here, we show the results of MLN-MRC according to different decision values 
(fusion weights). More analysis of different fusion strategies will be illustrated further in Section \ref{fus}.

The overall accuracy under different numbers of training samples 
per class (TS/C) is shown in Table~\ref{oa}. In this experiment, 
parameters of multiple resolutions are tuned for different HSIs. 
From the test results, the proposed MLN-MRC exhibits a superior overall performance, especially for those scenarios with fewer classes. 
MLN-SRC is marginally better than the SPCA. Note that, since our
MLN-based methods are easily integrable with 
various dimension reduction algorithms, we can combine
MLN-SRC with SPCA to further improve the performance. 
Because of page limitation, we leave the integration
of MLN-based methods and other dimension reduction algorithms in
future studies.

In real/practical scenarios, tuning parameters is always important for 
performance improvement. Since the prior knowledge of the optimal parameter 
choice is not always available, we also test the performance of MSPCA and MLN-MRC under the same set of input superpixel numbers to facilitate
fair comparison. In 
\ref{hsi}, we fuse the results from 9 resolutions, i.e., $N_i\in[25,35,50,70,100,140,200,280,400]$, for all comparative
methods. From these numerical results, we see that MLN-MRC provides an explicit improvement over MSPCA given fewer numbers of training samples since
MLN-based methods can better exploit the underlying correlation 
among all pixels. When given enough TS/C, MSPCA exhibits 
performance similar to that of MLN-MRC.

\begin{table}[t]
	\caption{Performance of Different Fusion Strategies}
	\label{fs}
	\begin{tabular}{|lllllll|}
		\hline
		\multicolumn{1}{|l|}{}                                                                    & \multicolumn{1}{l|}{TS/C} & \multicolumn{1}{l|}{MV}     & \multicolumn{1}{l|}{VA}              & \multicolumn{1}{l|}{DV}              & \multicolumn{1}{l|}{VN}              & TV              \\ \hline
		\multicolumn{7}{|l|}{Indian Pines}                                                                                                                                                                                                                                                         \\ \hline
		\multicolumn{1}{|l|}{\multirow{2}{*}{\begin{tabular}[c]{@{}l@{}}MS-\\ PCA\end{tabular}}}  & \multicolumn{1}{l|}{5}    & \multicolumn{1}{l|}{0.6663} & \multicolumn{1}{l|}{0.7241}          & \multicolumn{1}{l|}{\textbf{0.7334}} & \multicolumn{1}{l|}{0.7263}          & 0.7265          \\ \cline{2-7} 
		\multicolumn{1}{|l|}{}                                                                    & \multicolumn{1}{l|}{10}   & \multicolumn{1}{l|}{0.7476} & \multicolumn{1}{l|}{0.7778}          & \multicolumn{1}{l|}{\textbf{0.7829}} & \multicolumn{1}{l|}{0.7661}          & 0.7661          \\ \hline
		\multicolumn{1}{|l|}{\multirow{2}{*}{\begin{tabular}[c]{@{}l@{}}MLN-\\ MRC\end{tabular}}} & \multicolumn{1}{l|}{5}    & \multicolumn{1}{l|}{0.7164} & \multicolumn{1}{l|}{\textbf{0.7338}} & \multicolumn{1}{l|}{0.7260}          & \multicolumn{1}{l|}{0.7271}          & 0.7295          \\ \cline{2-7} 
		\multicolumn{1}{|l|}{}                                                                    & \multicolumn{1}{l|}{10}   & \multicolumn{1}{l|}{0.8220} & \multicolumn{1}{l|}{0.8253}          & \multicolumn{1}{l|}{\textbf{0.8253}} & \multicolumn{1}{l|}{0.7886}          & 0.7823          \\ \hline
		\multicolumn{7}{|l|}{Pavia University}                                                                                                                                                                                                                                                     \\ \hline
		\multicolumn{1}{|l|}{\multirow{2}{*}{\begin{tabular}[c]{@{}l@{}}MS-\\ PCA\end{tabular}}}  & \multicolumn{1}{l|}{5}    & \multicolumn{1}{l|}{0.7078} & \multicolumn{1}{l|}{0.7440}          & \multicolumn{1}{l|}{0.7317}          & \multicolumn{1}{l|}{\textbf{0.7798}} & 0.7712          \\ \cline{2-7} 
		\multicolumn{1}{|l|}{}                                                                    & \multicolumn{1}{l|}{10}   & \multicolumn{1}{l|}{0.8447} & \multicolumn{1}{l|}{0.8541}          & \multicolumn{1}{l|}{0.8553}          & \multicolumn{1}{l|}{\textbf{0.8588}} & 0.8500          \\ \hline
		\multicolumn{1}{|l|}{\multirow{2}{*}{\begin{tabular}[c]{@{}l@{}}MLN-\\ MRC\end{tabular}}} & \multicolumn{1}{l|}{5}    & \multicolumn{1}{l|}{0.8162} & \multicolumn{1}{l|}{0.8140}          & \multicolumn{1}{l|}{\textbf{0.8240}} & \multicolumn{1}{l|}{0.7977}          & 0.8085          \\ \cline{2-7} 
		\multicolumn{1}{|l|}{}                                                                    & \multicolumn{1}{l|}{10}   & \multicolumn{1}{l|}{0.8549} & \multicolumn{1}{l|}{0.8586}          & \multicolumn{1}{l|}{\textbf{0.8605}} & \multicolumn{1}{l|}{0.8459}          & 0.8435          \\ \hline
		\multicolumn{7}{|l|}{Salinas}                                                                                                                                                                                                                                                              \\ \hline
		\multicolumn{1}{|l|}{\multirow{2}{*}{\begin{tabular}[c]{@{}l@{}}MS-\\ PCA\end{tabular}}}  & \multicolumn{1}{l|}{5}    & \multicolumn{1}{l|}{0.8521} & \multicolumn{1}{l|}{0.9256}          & \multicolumn{1}{l|}{0.9349}          & \multicolumn{1}{l|}{\textbf{0.9655}} & 0.9634          \\ \cline{2-7} 
		\multicolumn{1}{|l|}{}                                                                    & \multicolumn{1}{l|}{10}   & \multicolumn{1}{l|}{0.9662} & \multicolumn{1}{l|}{0.9774}          & \multicolumn{1}{l|}{0.9718}          & \multicolumn{1}{l|}{\textbf{0.9789}} & 0.9765          \\ \hline
		\multicolumn{1}{|l|}{\multirow{2}{*}{\begin{tabular}[c]{@{}l@{}}MLN-\\ MRC\end{tabular}}} & \multicolumn{1}{l|}{5}    & \multicolumn{1}{l|}{0.9469} & \multicolumn{1}{l|}{0.9456}          & \multicolumn{1}{l|}{0.9485}          & \multicolumn{1}{l|}{\textbf{0.9844}} & 0.9622          \\ \cline{2-7} 
		\multicolumn{1}{|l|}{}                                                                    & \multicolumn{1}{l|}{10}   & \multicolumn{1}{l|}{0.9872} & \multicolumn{1}{l|}{0.9874}          & \multicolumn{1}{l|}{0.9873}          & \multicolumn{1}{l|}{0.9857}          & \textbf{0.9942} \\ \hline
	\end{tabular}
\end{table}

\begin{figure*}[t]
	\centering
	\subfigure[VN-based Weights]{
		\label{F1}
		\includegraphics[height=4cm]{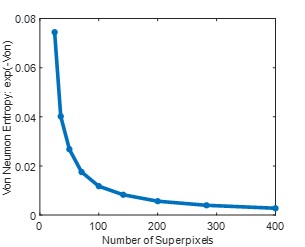}}
	\hfill
	\subfigure[TV-based Weights]{
		\label{F2}
		\includegraphics[height=4cm]{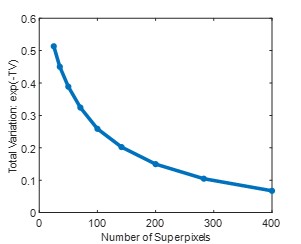}}
	\hfill
	\subfigure[VA in M-GSP]{
		\label{F3}
		\includegraphics[height=4cm]{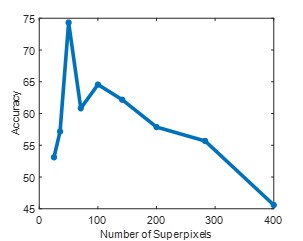}}\\
	\subfigure[VA in MSPCA]{
		\label{F4}
		\includegraphics[height=4cm]{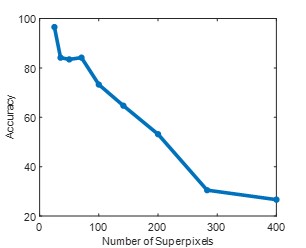}}
	\hfill
	\subfigure[Mean DV in M-GSP]{
		\label{F5}
		\includegraphics[height=4cm]{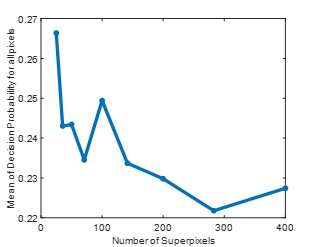}}
	\hfill
	\subfigure[Mean DV in MSPCA]{
		\label{F6}
		\includegraphics[height=4cm]{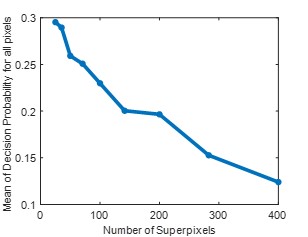}}
	\caption{Decision Strengths over Different Numbers of Superpixels.}
	\label{FF}
\end{figure*}
\begin{table*}[t]
	\centering
	\caption{Overall Accuracy in Noisy Environment}
	\label{ovnoise}
	\begin{tabular}{|lllllllllll|}
		\hline
		\multicolumn{2}{|l|}{\textbf{(Uniform/Gaussian)}}                            & \multicolumn{3}{l|}{Indian Pines}                                                & \multicolumn{3}{l|}{Pavia University}                                                     & \multicolumn{3}{l|}{Salinas}                                         \\ \hline
		\multicolumn{1}{|l|}{TS/C}                & \multicolumn{1}{l|}{Noise Level} & \multicolumn{2}{l|}{MPCA}          & \multicolumn{1}{l|}{MLN-MRC}                & \multicolumn{2}{l|}{MSPCA}                  & \multicolumn{1}{l|}{MLN-MRC}                & \multicolumn{2}{l|}{MSPCA}                  & MLN-MRC                \\ \hline
		\multicolumn{11}{|l|}{Setup 1}                                                                                                                                                                                                                                                                                                     \\ \hline
		\multicolumn{1}{|l|}{\multirow{3}{*}{5}}  & \multicolumn{1}{l|}{5\%}         & \multicolumn{2}{l|}{0.7140/0.7012} & \multicolumn{1}{l|}{\textbf{0.7377/0.7493}} & \multicolumn{2}{l|}{0.7219/0.7160}          & \multicolumn{1}{l|}{\textbf{0.8230/0.8126}} & \multicolumn{2}{l|}{0.8592/0.8670}          & \textbf{0.9231/0.9159} \\ \cline{2-11} 
		\multicolumn{1}{|l|}{}                    & \multicolumn{1}{l|}{10\%}        & \multicolumn{2}{l|}{0.7110/0.6702} & \multicolumn{1}{l|}{\textbf{0.7135/0.7276}} & \multicolumn{2}{l|}{0.7080/0.7031}          & \multicolumn{1}{l|}{\textbf{0.8085/0.7873}} & \multicolumn{2}{l|}{0.8341/0.8421}          & \textbf{0.8956/0.9023} \\ \cline{2-11} 
		\multicolumn{1}{|l|}{}                    & \multicolumn{1}{l|}{15\%}        & \multicolumn{2}{l|}{0.6868/0.6507} & \multicolumn{1}{l|}{\textbf{0.7133/0.7215}} & \multicolumn{2}{l|}{0.6831/0.6817}          & \multicolumn{1}{l|}{\textbf{0.7932/0.7640}} & \multicolumn{2}{l|}{0.8107/0.8215}          & \textbf{0.8815/0.8975} \\ \hline
		\multicolumn{1}{|l|}{\multirow{3}{*}{10}} & \multicolumn{1}{l|}{5\%}         & \multicolumn{2}{l|}{0.7613/0.7753} & \multicolumn{1}{l|}{\textbf{0.8033/0.8056}} & \multicolumn{2}{l|}{\textbf{0.8499}/0.8374} & \multicolumn{1}{l|}{0.8414/\textbf{0.8475}}          & \multicolumn{2}{l|}{0.9509/0.9647}          & \textbf{0.9632/0.9684} \\ \cline{2-11} 
		\multicolumn{1}{|l|}{}                    & \multicolumn{1}{l|}{10\%}        & \multicolumn{2}{l|}{0.7522/0.7498} & \multicolumn{1}{l|}{\textbf{0.7788/0.7984}} & \multicolumn{2}{l|}{0.8481/0.8436}          & \multicolumn{1}{l|}{\textbf{0.8597/0.8461}} & \multicolumn{2}{l|}{0.9408/0.9431}          & \textbf{0.9445/0.9491} \\ \cline{2-11} 
		\multicolumn{1}{|l|}{}                    & \multicolumn{1}{l|}{15\%}        & \multicolumn{2}{l|}{0.7317/0.7382} & \multicolumn{1}{l|}{\textbf{0.7544/0.7845}} & \multicolumn{2}{l|}{0.8350/0.7848}          & \multicolumn{1}{l|}{\textbf{0.8467/0.8333}} & \multicolumn{2}{l|}{0.9381/\textbf{0.9368}} & \textbf{0.9421}/0.9317          \\ \hline
		\multicolumn{11}{|l|}{Setup 2}                                                                                                                                                                                                                                                                                                     \\ \hline
		\multicolumn{1}{|l|}{\multirow{3}{*}{5}}  & \multicolumn{1}{l|}{5\%}         & \multicolumn{2}{l|}{0.6839/0.6740} & \multicolumn{1}{l|}{\textbf{0.7023/0.6898}} & \multicolumn{2}{l|}{0.7328/0.7093}          & \multicolumn{1}{l|}{\textbf{0.8422/0.8390}} & \multicolumn{2}{l|}{0.8591/0.8679}          & \textbf{0.9240/0.9050} \\ \cline{2-11} 
		\multicolumn{1}{|l|}{}                    & \multicolumn{1}{l|}{10\%}        & \multicolumn{2}{l|}{0.6640/0.6501} & \multicolumn{1}{l|}{\textbf{0.6764/0.6647}} & \multicolumn{2}{l|}{0.7052/0.6972}          & \multicolumn{1}{l|}{\textbf{0.8023/0.7837}} & \multicolumn{2}{l|}{0.8471/0.8512}          & \textbf{0.9045/0.8878} \\ \cline{2-11} 
		\multicolumn{1}{|l|}{}                    & \multicolumn{1}{l|}{15\%}        & \multicolumn{2}{l|}{0.6413/0.6247} & \multicolumn{1}{l|}{\textbf{0.6687/0.6427}} & \multicolumn{2}{l|}{0.6754/0.6789}          & \multicolumn{1}{l|}{\textbf{0.7635/0.7530}} & \multicolumn{2}{l|}{0.8317/0.8421}          & \textbf{0.8915/0.8775} \\ \hline
		\multicolumn{1}{|l|}{\multirow{3}{*}{10}} & \multicolumn{1}{l|}{5\%}         & \multicolumn{2}{l|}{0.7579/0.7408} & \multicolumn{1}{l|}{\textbf{0.8022/0.7810}} & \multicolumn{2}{l|}{\textbf{0.8649}/0.8437} & \multicolumn{1}{l|}{0.8573/\textbf{0.8632}} & \multicolumn{2}{l|}{0.9682/0.9621}          & \textbf{0.9735/0.9633} \\ \cline{2-11} 
		\multicolumn{1}{|l|}{}                    & \multicolumn{1}{l|}{10\%}        & \multicolumn{2}{l|}{0.7409/0.7333} & \multicolumn{1}{l|}{\textbf{0.7800/0.7762}} & \multicolumn{2}{l|}{0.8442/0.8354}          & \multicolumn{1}{l|}{\textbf{0.8469/0.8501}} & \multicolumn{2}{l|}{0.9521/0.9450}          & \textbf{0.9547/0.9532} \\ \cline{2-11} 
		\multicolumn{1}{|l|}{}                    & \multicolumn{1}{l|}{15\%}        & \multicolumn{2}{l|}{0.7208/0.7235} & \multicolumn{1}{l|}{\textbf{0.7695/0.7584}} & \multicolumn{2}{l|}{\textbf{0.8410/0.8309}} & \multicolumn{1}{l|}{0.8368/0.8258}          & \multicolumn{2}{l|}{\textbf{0.9437/0.9416}} & 0.9305/0.9241          \\ \hline
	\end{tabular}
\end{table*}
\subsubsection{Analysis of Different Fusion Strategies}\label{fus}
Now, we analyze the performance of different fusion strategies. Since 
MSPCA \cite{c15} also fuses multiple resolutions of superpixels, 
we investigate our proposed fusion methods in 
combination with both MSPCA and MLN-MRC.

We follow the same setup of $9$ resolution as Table \ref{hsi}. The test accuracy is shown in Fig. \ref{FF}. As shown, our proposed weight approaches
lead to significant improvement over the basic majority voting (VT) for MSPCA. Compared to SVM-based weights, these graph-based weights are better for MSPCA since additional geometric information is considered. For MLN-MRC, the proposed weights show a slight improvement, which suggests that
MLN-MRC is less sensitive to different decision strengths. 
To better understand the effect of different weights, we 
also present decision strengths over different numbers of superpixels in 
Fig.~\ref{FF}. In Figs.~\ref{F1}-\ref{F2}, the graph-based weights favor low resolution to form baselines and use high resolutions to interpolate details. 
Thus, for HSIs with larger area of segmented groups, such as \textit{Salinas}, the graph-based weights generate superior performances. 
Using SVM-based weights, we find no consistent trend among
different superpixels, and the results vary for different datasets. Since the MLN-based methods have already incorporated the underlying geometric structures, 
they continue to display robust results even when using SVM-based weights.

\subsubsection{Robustness in Noisy Datasets}
We further evaluate the robustness of proposed methods in noisy environment. More specifically, we consider two types of noise models: 1) pixel-dependent noise 
where pixel noise variance depending on corresponding
pixel data value; and 2) non-pixel dependent noise where noise variance
is defined by mean of all pixel values. These two different noise models
describe two different practical sensing noises. 
We also test both uniform noise and Gaussian noise. From
the test performances shown in Table~\ref{ovnoise}, we find the 
newly proposed MLN methods to be less sensitive to various
types of sample noises.

\subsubsection{Complexity}
Since our proposed M-GSP processing can be flexibly integrated with other dimension reduction methods to reduce complexity, we find it unnecessary to provide
evaluation of computation complexity for various setups. In general, the original 
MLN-MRC has a similar runtime as MSPCA. For example, under 
the same settings of multi-resolutions as Section \ref{oac} in \textit{Indian Pines} HSI with tuning parameters of SVM among 15 sets, 
the runtimes for MLN-MRC and MSPCA are $28.21$ seconds and $25.37$ seconds, respectively. These and other tests indicate
similar computation complexity for methods based on M-GSP and PCA.

\section{Conclusions}\label{con}
This work introduces the use of M-GSP in hyperspectral imaging processing. To capture heterogeneous underlying structures within different but 
highly correlated spectrum frames in hyperspectral images (HSIs), 
we propose to represent HSIs via multilayer graph networks. 
Analyzing singular spectra of 
adjacency tensor for the multilayer network,
we first develop a MLN-based spectral clustering for unsupervised HSI 
segmentation. Extracting features based on MLN-GSP, we then propose 
two algorithms for the semi-supervised HSI segmentation.
We also consider several novel decision fusion strategies for
multiple resolution superpixel analysis. Our
experimental results demonstrate the robustness and efficiency of 
the proposed methods, successfully showcasing the power of M-GSP in HSI analysis.

Since the proposed MLN-based methods is amenable to integration with 
various dimension reduction and feature extraction methods, 
future works may consider M-GSP in connection with other advanced HSI processing approaches for performance improvement and for robustness. 
Another interesting future direction is to extract more 
valuable features via MLN Fourier transform (M-GFT) beyond only spectrum decomposition. Additionally, we also find the challenge of 
processing unlabeled samples (background) an exciting future research
direction.

\end{document}